\pdfoutput=1

\documentclass[11pt]{article}

\usepackage{arr}

\usepackage{times}
\usepackage{latexsym}

\usepackage{multirow}
\usepackage{hhline}
\usepackage{amsmath,amsfonts,amssymb}
\usepackage{pgfplots}
\pgfplotsset{compat=1.9}
\usepackage{graphicx}
\usepackage{float}
\usepackage{bm}
\usepackage{soul}
\usepackage{verbatim}
\usepackage{booktabs, cellspace}
\usepackage{caption}
\usepackage[inline]{enumitem}
\usepackage[final]{pdfpages}
\usepackage[T1]{fontenc}
\usepackage{subcaption}
\usepackage{wrapfig}
\usepackage{array}
\usepackage{color, colortbl}
\usepackage{placeins}

\usepackage{caption}

\newlist{inlineenum}{enumerate*}{1}
\setlist[inlineenum]{label=(\roman*)}

\usepackage[textsize=tiny,disable]{todonotes}

\usepackage[T1]{fontenc}

\usepackage[utf8]{inputenc}

\usepackage{microtype}

\title{When Does Monolingual Data Help Multilingual Translation: \\The Role of Domain and Model Scale}

\author{
    \hspace{-2em}Christos Baziotis\thanks{\quad Work done while at University of Edinburgh.}  \\
    {\hspace{-2em}Samaya AI} \\
    {\hspace{-2em}\small \texttt{christos@samaya.ai}}
    \And
    \hspace{-5em}Biao Zhang\footnotemark[1] \\
    {\hspace{-5em}Google DeepMind} \\
    {\hspace{-5em}\small \texttt{biaojiaxing@google.com}}
    \And
    \hspace{-5em}Alexandra Birch \quad Barry Haddow \\
    {\hspace{-5em}University of Edinburgh} \\
    {\hspace{-5em}\small \texttt{\{a.birch, bhaddow\}@ed.ac.uk}}
}

\begin{document}
    \maketitle
    \begin{abstract}

        Multilingual machine translation (MMT), trained on a mixture of parallel and monolingual data,
        is key for improving translation in low-resource language pairs.
        However, the literature offers
        conflicting results on the performance of different methods of including monolingual data.
        To resolve this,
        we examine how denoising autoencoding (DAE) and backtranslation (BT) impact MMT
        under different data conditions and model scales.
        Unlike prior studies, we use a realistic dataset of 100 translation directions
        and consider many domain combinations of monolingual and test data.
        We find that monolingual data generally helps MMT,
        but models are surprisingly brittle to domain mismatches,
        especially at smaller model scales.
        BT is beneficial when the parallel, monolingual, and test data sources are similar but can be detrimental otherwise, while DAE is less effective than previously reported.
        Next, we analyze the impact of scale (from 90M to 1.6B parameters) and find it is important for both methods, particularly DAE.
        As scale increases, DAE transitions from underperforming the parallel-only baseline at 90M to converging with BT performance at 1.6B, and even surpassing it in low-resource.
        These results offer new insights into how to best use monolingual data in MMT.

    \end{abstract}

    \section{Introduction}

    The need for large supervised corpora remains a major bottleneck
    in neural machine translation (NMT)~\cite{bapna2022building}.
    Sufficient bilingual data is scarce for most languages and
    limited to religious texts for the lowest-resource languages.
    To compensate for this lack of data,
    one effective approach is to leverage \textit{related parallel data} from other languages via
    multilingual machine translation (MMT) that enables positive transfer from high-resource
    to low-resource languages~\cite{aharoni-etal-2019-massively,arivazhagan2019massively}.
    Additionally, we can use \textit{monolingual data},
    either through pretraining with denoising autoencoding(DAE;~\citealt{Conneau2019-fk,liu-etal-2020-multilingual-denoising}),
    or with backtranslation (BT;~\citealp{sennrich-etal-2016-improving}).
    Driven by the success of these methods,
    recent works are converging toward a unified approach,
    that jointly trains MMT with monolingual data using auxiliary
    DAE objectives~\cite{siddhant2022towards,bapna2022building,nllb2022} and/or BT.

    However, the literature contains contradictory results about the effectiveness of these methods, particularly DAE.
    Early studies indicated combining MMT with DAE led to improvements across all settings~\cite{wang-etal-2020-multi, siddhant-etal-2020-leveraging}.
    These studies, however, were limited in scope, as they only considered moderately-sized models and used few languages (10 to 15), with training and test data drawn from similar domains.
    By contrast,
    \citet{nllb2022} found that DAE helped only in very low-resource directions
    in MMT experiments with 200+ languages,
    while \citet{xu2023languageaware} reported that DAE produced mixed results in experiments with (mostly) African languages.

    To resolve this conflict,
    we present a systematic analysis of different methods that integrate monolingual data into MMT,
    focusing on BT and two DAE objectives, MASS~\cite{song2019-dc} and BART~\cite{lewis-etal-2020-bart, Liu2020-un}.
    First, we carefully investigate the role of the \textit{domain}.
    To align with prior work, we focus on the English-centric setting (i.e., concatenation of English$\rightarrow$XX and XX$\rightarrow$English).
    We use a realistic and diverse multilingual translation dataset with 100 directions and run controlled experiments using different monolingual splits with single- and mixed-domain data.
    Then, we evaluate models across four wide-coverage multilingual test sets from Wikipedia, news, medical, and mixed domains.
    Our results with medium-sized models (370M)
    show that while BT outperforms both DAE objectives in most settings,
    the effectiveness of all methods varies significantly, as they are surprisingly brittle to domain mismatches.
    BT is more sensitive to the domain than DAE, and can underperform the parallel-only baseline
    when the monolingual and test data are not similar.
    However, increasing the diversity of the monolingual data by mixing different sources
    improves domain robustness to some extent.
    We also discover that both DAE methods are less effective than previously reported,
    and they are mainly helpful in low-resource and xx$\rightarrow$en directions.
    Of the two, MASS consistently outperforms BART, although by a narrow margin.

    Next, we study the role of \textit{model capacity}
    and discover that it is crucial and can even change the ranking between methods.
    We hold all other factors constant and train models
    with sizes from 90M up to 1.6B parameters.
    When the scale is small,
    both BT and DAE yield poor results,
    especially in out-of-domain settings.
    However, as model capacity grows, all methods quickly improve compared to the parallel-only baseline,
    and also become more robust to domain mismatches.
    Scale affects DAE the most, which transitions from underperforming the parallel-only baseline at the 90M scale
    to becoming competitive with BT at 1.6B and even outperforming it in low-resource.

    Our contributions are:
    \begin{inlineenum}
        \item We present a large-scale systematic analysis of how the \textit{domain} and model \textit{scale} affect the effectiveness of methods that incorporate monolingual data into MMT.
        \item We show that BT and DAE are sensitive to domain mismatches between the monolingual and test data,
        particularly on small scales.
        BT is best in most settings.
        Also, prior works have overestimated DAE, and when comparing the two methods, MASS outperforms BART.
        \item We discover that model capacity is key for the effectiveness of both methods, especially DAE. When the scale is small, DAE can even harm MMT,
        but it quickly improves with scale, and eventually becomes competitive with BT.
    \end{inlineenum}

    \section{Related Work}
    \label{sec:related-work}

    \paragraph{Monolingual Data with Multi-Task Learning}

    Early works on DAE+MMT report universal gains in all settings.
    \citet{siddhant-etal-2020-leveraging}
    use WMT parallel data from 15 languages and large monolingual corpora from many sources,
    like News Crawl, Wikipedia, and Common Crawl, with MASS.
    \citet{wang-etal-2020-multi} explore BART-like objectives with a subset of 10 languages from \citet{siddhant-etal-2020-leveraging} and News Crawl monolingual data.

    However, more recent works that use larger and/or less uniform datasets, report less favourable results.
    To extend MMT to very low-resource languages,
    \citet{bapna2022building} show that models learn to translate from/into languages with only monolingual data
    if there are sufficient parallel data in other languages to enable transfer from the DAE to the MT task.
    \citet{nllb2022} explore a similar idea, but report that, in supervised translation,
    DAE (BART) is effective only for very low-resource.
    \citet{xu2023languageaware} compare all aforementioned
    DAE methods and find that they often fail to outperform the parallel-only baseline.
    Our study probes confounding factors in these prior works.

    \paragraph{Large Language Models}
    Large language models (LLMs) trained on massive datasets achieve impressive results in many tasks~\cite{brown2020language, chowdhery2022palm, zhang2022opt, tay2023ul}.
    To adapt LLMs to downstream tasks including translation~\cite{weiemergent, lin-etal-2022-shot, zhang2023prompting, vilar2022prompting, garcia2023unreasonable, zhu2023multilingual, hendy2023good},
    the dominant approach is to use prompting, an ability enabled by model scale~\cite{weiemergent}.
    Our work, however, is orthogonal and presents an analysis of methods that integrate monolingual data into encoder-decoder MMT models trained from scratch.
    Also, it is questionable whether these models are unsupervised with respect to translation,
    as recent work suggests that they have consumed parallel data during pretraining~\cite{briakou2023searching}.

    \paragraph{Model Scale}
    A growing literature investigates the scaling laws of different aspects of a model~\cite{kaplan2020scaling}.
    In NMT, \citet{ghorbani2021scaling} explore scaling laws related to model capacity,
    \citet{fernandes2023scaling} consider MMT,
    and \citet{gordon-etal-2021-data} focus on data scaling.
    \citet{zhang2022examining} investigate the scaling laws across architectures,
    like decoder-only and encoder-decoder.
    Our work does not study scaling laws but analyzes how scale impacts using monolingual data in MMT.

    \paragraph{Analysis}
    \citet{huang-etal-2021-comparison,liu-etal-2021-complementarity-pre}
    analyze the complementarity of BT and monolingual pretraining when used in bilingual NMT.
    By contrast, we focus on multilingual NMT and systematically analyze the joint training with BT and DAE.

    \section{(Multi-task) Multilingual NMT}

    \label{sec:mnmt}
    We follow the universal MMT training method of \citet{johnson2017multilingual} and
    train a single dense Transformer-based~\cite{vaswani2017attention} model on the concatenation of parallel data
    from multiple language pairs.
    We prepend a special token $\langle2\textsc{xx}\rangle$ to the source sequences,
    that informs the model about the translation direction (e.g., $\langle2\textsc{es}\rangle$ for Spanish).

    \subsection{Denoising Autoencoding}

    We follow the multi-task setting from prior works~\cite{siddhant-etal-2020-leveraging,wang-etal-2020-multi}
    and use the regular MT objective on batches with parallel data
    and a DAE objective on batches with monolingual data.
    The language token $\langle2\textsc{xx}\rangle$ informs the model about the DAE and MT tasks,
    as it instructs it to generate a semantically similar sentence in the \textsc{xx} language.
    We explore two DAE methods.

    \begin{figure}[t]
        \centering
        \includegraphics[width=1\columnwidth]{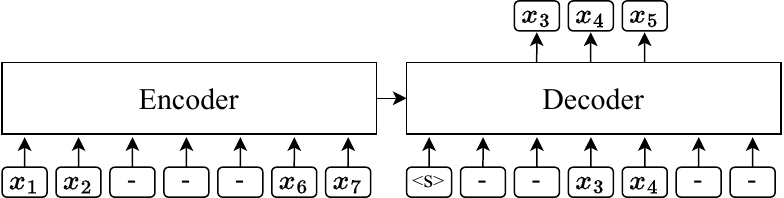}
        \caption{Illustration of the MASS objective.}
        \label{fig:mass}
        \vspace{4pt}
    \end{figure}
    \begin{figure}[t]
        \centering
        \includegraphics[width=1\columnwidth]{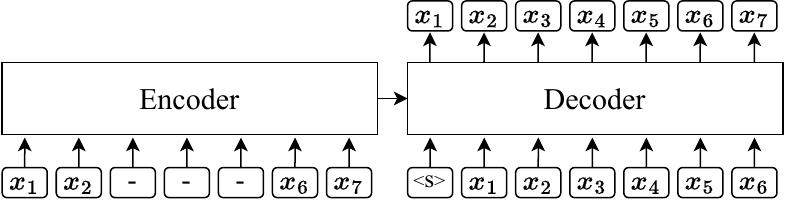}
        \caption{Illustration of the BART objective.}
        \label{fig:bart}
    \end{figure}

    \smallskip\noindent\textbf{MASS}
    \citet{song2019-dc} adapt the masked language modeling objective~\cite{devlin-etal-2019-bert}
    to encoder-decoder models.
    MASS masks a span in the input and
    trains the decoder to predict that span.
    However, the unmasked tokens are not included in the target prefix (Figure~\ref{fig:mass}).
    Following \citet{siddhant-etal-2020-leveraging, siddhant2022towards},
    we do not use the architectural modifications of \citet{song2019-dc},
    such as extra language embeddings or custom initialization.

    \smallskip
    \noindent\textbf{BART}
    \citet{lewis-etal-2020-bart} propose a DAE objective similar to MASS, but with two differences.
    First, BART uses a slightly different noising strategy that can corrupt more than one input span in each sentence.
    Second, and more importantly, while the decoder is also trained to reconstruct the source sentence,
    its input context contains the full prefix, \textit{including} the masked tokens (Figure~\ref{fig:bart}).

    \subsection{Backtranslation}
    \label{sec:mnmt-backtranslation}

    For BT, to save resources, instead of training separate bilingual models,
    we re-use the baseline MMT model and generate the new synthetic parallel data using the monolingual data of each language.

    \section{Experimental Setup}

    \smallskip\noindent\textbf{Parallel Data}
    We use ML50~\cite{tang-etal-2021-multilingual},
    a multilingual translation dataset between English and 50 other languages.
    ML50 is more representative of real-world multilingual datasets as it contains typologically diverse languages,
    including high, medium, and (extremely -- less than 10k) low resource pairs, and with data from different
    domains.
    It is also more multilingual than the datasets from \citet{siddhant-etal-2020-leveraging}
    and \citet{wang-etal-2020-multi}, that use 15 and 10 languages, respectively.
    To reduce training time, we cap the parallel data at 10M sentences per language,
    similar to~\citealt{wang-etal-2020-multi}, which affects only few high-resource languages.

    \smallskip\noindent\textbf{Monolingual Data}
    We run \textit{controlled} experiments with single- and mixed-domain monolingual data.
    For the single-domain experiments,
    we use Wikipedia as it is the only publicly available source
    with available data for all languages in ML50,
    but exclude the \textit{xh} and \textit{iu} languages from the experiments
    as they lack sufficient monolingual data.
    We cap the monolingual data per language to 5M,
    similar to \citet{wang-etal-2020-multi},
    which is still much larger than the parallel data for most languages.
    For the mixed-domain experiments, we use the \textit{same} number of sentences per language,
    but also include News Crawl\footnotemark
    \footnotetext{\href{www.statmt.org/wmt20/translation-task.html}{https://www.statmt.org/wmt20/translation-task.html}} \citep{barrault-etal-2020-findings}
    and Web Crawl data from CC100\footnotemark~\citep{conneau-etal-2020-unsupervised}.
    \footnotetext{\href{https://data.statmt.org/cc-100/}{https://data.statmt.org/cc-100/}}
    See the Appendix for the full data statistics (Table~\ref{table:full-data-stats}).

    \smallskip\noindent\textbf{Evaluation}
    Besides ML50 we also consider three \textit{domain-specific} test sets.
    We use FLORES-200 \cite{goyal-etal-2022-flores, nllb2022} with translations of \textit{Wikipedia} articles,
    NTREX-128\footnotemark~\cite{federmann-etal-2022-ntrex} with translations in 128 languages
    from the English WMT19 \textit{News} test set~\cite{barrault-etal-2019-findings},
    \footnotetext{\href{https://github.com/MicrosoftTranslator/NTREX}{https://github.com/MicrosoftTranslator/NTREX}.
    Because of misalignments, we omit the \textit{ur} and \textit{vi} languages.}
    and TICO-19 with translations in the \textit{medical} domain~\cite{anastasopoulos-etal-2020-tico}.
    FLORES-200 and NTREX-128 cover all languages in ML50, while TICO-19 covers only 15,
    but equally distributed across high, medium, and low resources.
    At test time, use beam search with K=5.
    In the main paper,
    we report results using BLEU~\cite{papineni-etal-2002-bleu} similar to most prior works.
    However, to make our evaluation more comprehensive,
    we include in the Appendix the results from all experiments using
    ChrF~\cite{popovic-2015-chrf} and COMET\footnotemark~\cite{rei-etal-2020-comet},
    which is a neural metric.
    \footnotetext{We use v2.0.1 with the \textit{wmt22-comet-da} model.}
    We find that overall, all metrics are very consistent with each other,
    with few small differences in en$\rightarrow$xx (see Appendix).
    We use SacreBLEU\footnotemark~\cite{post-2018-call} for ChrF and BLEU.
    \footnotetext{\textls[-60]{\fontsize{8.5}{8}\selectfont  BLEU+case.mixed+lang.S-T+numrefs.1+smooth.exp+tok.13a+v1.5.1}}

    \begingroup
\setlength{\tabcolsep}{3.7pt} %
\renewcommand{\arraystretch}{1.1} %
\begin{table}
	\small
	\centering

	\begin{tabular}{lrrrrrrr}
		\toprule[1.5pt]
		\multirow{2}[3]{*}{Model} &
		\multicolumn{3}{c}{en$\rightarrow$xx} &
		\multicolumn{3}{c}{xx$\rightarrow$en} &
		\multicolumn{1}{c}{\multirow{2}[3]{*}{Mean}}
		\\
		\cmidrule(lr){2-4} \cmidrule(lr){5-7}
		&
		High & Med & Low &
		High & Med & Low &
		\\
		\midrule
				
		parallel & 22.5                   & 17.3                   & 20.6          & 26.9          & 25.9          & 25.3          & 22.9          \\
		+BART    & \cellcolor{red!15}22.0 & \cellcolor{red!15}16.9 & 21.3          & 27.0          & 26.6          & 27.9          & 23.6          \\
		+MASS    & \cellcolor{red!15}22.1 & \cellcolor{red!15}16.9 & 21.3          & 27.1          & 26.5          & 28.5          & 23.7          \\
		+BT      & \textbf{23.6}          & \textbf{17.3}          & \textbf{21.6} & \textbf{27.8} & \textbf{26.9} & \textbf{28.9} & \textbf{24.3} \\
		\bottomrule[1.5pt]
				
	\end{tabular}
	\caption{BLEU scores ($\uparrow$) on the ML50 test. The models with BT and both SSL objectives (BART, MASS) use
		the single-domain monolingual split with data only from Wikipedia.
		The cells in \colorbox{red!15}{red}  indicate that a model fails to improve over the parallel-only baseline.}
	\label{tab:mmt-wiki}
\end{table}
\endgroup

    \begin{figure*}[t]
        \centering
        \includegraphics[width=0.99\textwidth]{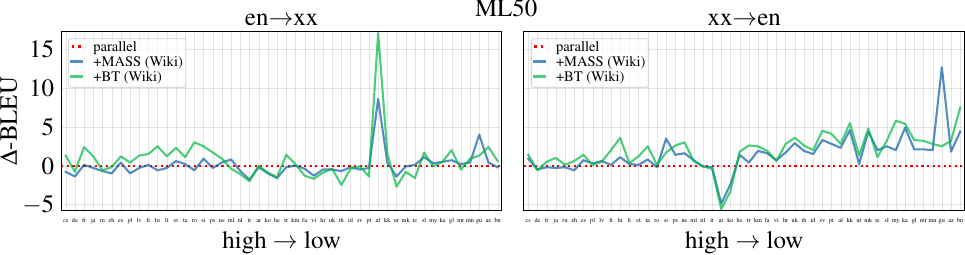}
        \caption{BLEU differences between each model and the parallel-only model (red dotted line) on the ML50 test data.}
        \label{fig:delta-bleu-wiki}
    \end{figure*}

    \smallskip\noindent\textbf{Data Sampling}
    We use temperature-based data sampling~\citep{arivazhagan2019massively} to balance the training data.
    Assuming that $p_\textsc{d}$ is the probability that a sentence belongs to dataset $D$,
    we sample sentences for $D$ with a probability proportional to $p_\textsc{d}^{1/T}$,
    where $T$ is a temperature parameter.
    When using parallel data, $D$ corresponds to the data of a given language pair.
    When including monolingual (i.e., for DAE) or synthetic parallel (i.e., for BT) data,
    we \textit{first} concatenate all the separate datasets to the same list and \textit{then} apply temperature sampling.
    That is, the real en$\rightarrow$fr, synthetic (BT)  en$'\rightarrow$fr, and monolingual fr$\leftrightarrow$fr,
    are treated as separate datasets $D$.
    Larger values of $T$ lead to more even sampling (i.e., upsampling small datasets).
    We set $T=5$ following prior works~\cite{wang-etal-2020-multi, siddhant-etal-2020-leveraging},
    which also leads to a roughly 1:1 ratio when using both monolingual and parallel data.

    \smallskip\noindent\textbf{Models}
    Our baseline is an MMT model trained \textit{only} on the en$\rightarrow$xx and xx$\rightarrow$en parallel data.
    For both MASS and BART, we mask 50\% of input tokens following the hyperparameters from \citet{siddhant2022towards,siddhant-etal-2020-leveraging}
    and \citet{nllb2022}, respectively.
    All models use the same Transformer architecture~\cite{vaswani2017attention}.
    We consider three different model sizes for our scaling experiments:
    1) \textit{Transformer-Base} with 90M parameters,
    2) \textit{Transformer-Big} with 370M parameters,
    and 3) \textit{Transformer-XL} (not to be confused with \citealt{dai-etal-2019-transformer}),
    with 1.6B parameters.
    We include details about our models and training in Appendix~\ref{sec:app-exp-setup}.

    \section{Results}
    \label{sec:results}

    \subsection{Single-Domain Monolingual Data (Wiki)}
    \label{sec:results-wiki-mono}

    We begin with a series of controlled experiments that measure the impact of the domain using the Transformer-Big model scale (370M).
    We compare across different test sets
    the parallel-only model with parallel+BT and parallel+DAE (MASS, BART) that use the single-domain monolingual split (see statistics in Table~\ref{table:full-data-stats}).
    In Table~\ref{tab:mmt-wiki} we report the BLEU scores of each model on the ML50 test set
    averaged by group and translation direction.

    On average, BT and both DAE models outperform the baseline
    by +1.4 and +0.7 BLEU points, respectively.
    BT consistently achieves the best results, with the largest gains in low-resource,
    with +1 BLEU points on en$\rightarrow$xx and +3.6 BLEU points on xx$\rightarrow$en.
    Both DAE models produce similar results, but MASS is marginally better.
    However, in the en$\rightarrow$xx high- and medium-resource languages, both DAE models fail to outperform the baseline,
    although they use the same monolingual data as BT.

    \begin{figure*}[t]
        \begin{subfigure}{\linewidth}
            \centering
            \includegraphics[width=0.97\textwidth]{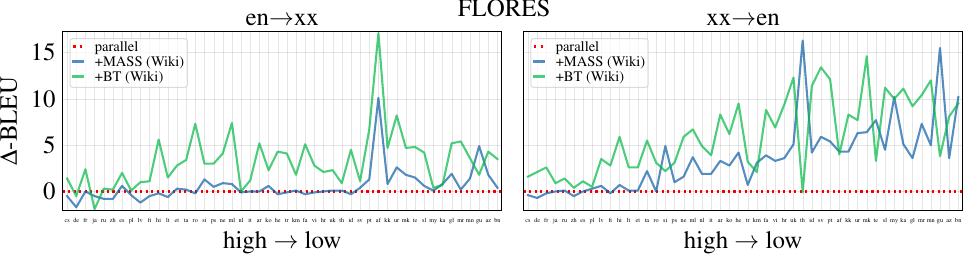}
        \end{subfigure}
        \hfill\par\vspace*{-10pt}
        \begin{subfigure}{\linewidth}
            \centering
            \includegraphics[width=0.97\textwidth]{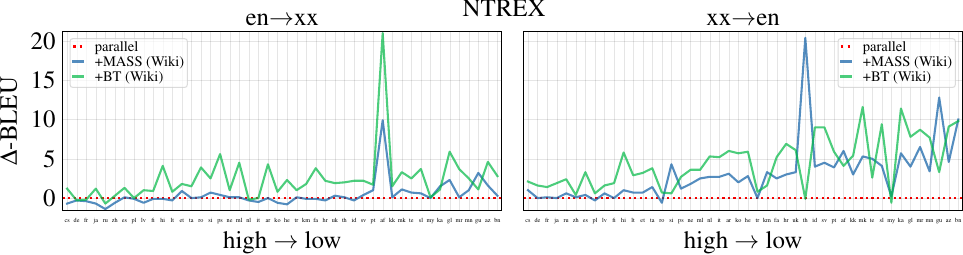}
        \end{subfigure}
        \caption{BLEU differences between each model and the baseline (red dotted line) on FLORES and NTREX.}
        \label{fig:delta-bleu-wiki-flores-ntrex}
    \end{figure*}

    \begin{figure}[t]
        \centering
        \includegraphics[width=1\linewidth]{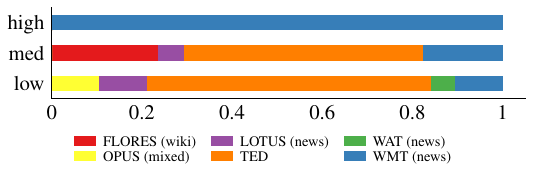}
        \captionof{figure}{Data sources used for the ML50 test sets.}
        \label{fig:ml50-data}
    \end{figure}

    \paragraph{Non-aggregated scores reveal mixed results.}
    To get a more detailed picture of model performance we plot the \textit{differences}
    in the BLEU scores ($\Delta$-BLEU) between each model and the parallel-only baseline
    model across all pairs in Figure~\ref{fig:delta-bleu-wiki}.
    For a simpler presentation, we omit BART which is similar to MASS.
    Figure~\ref{fig:delta-bleu-wiki} reveals that the results are more mixed than the aggregated scores suggest
    (Table~\ref{tab:mmt-wiki}).
    In xx$\rightarrow$en, both BT and MASS are generally better than the baseline and follow a similar trend.
    Their gains increase towards the low-resource languages, with few exceptions,
    and BT is better than MASS in most cases.
    However, in en$\rightarrow$xx, we discover a different picture.
    BT shows a surprising behavior as it outperforms the baseline in high-resource
    (usually from +2 to +4 BLEU)
    but harms BLEU in most medium- to low-resource languages
    and is also often worse than MASS.
    MASS fluctuates around the baseline and benefits only a few low-resource languages.
    These results contradict early works on MMT+DAE that report \textit{universal} gains~\cite{siddhant-etal-2020-leveraging,wang-etal-2020-multi}.

    \begin{figure*}[t]
        \begin{subfigure}{0.47\textwidth}
            \centering
            \includegraphics[width=\linewidth]{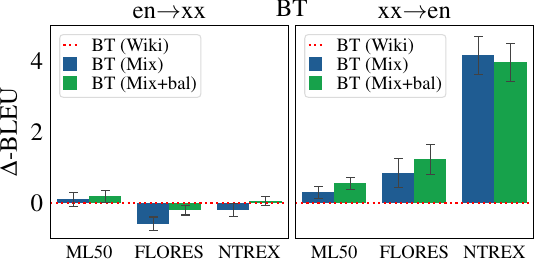}
        \end{subfigure}
        \hfil
        \begin{subfigure}{0.47\textwidth}
            \centering
            \includegraphics[width=\linewidth]{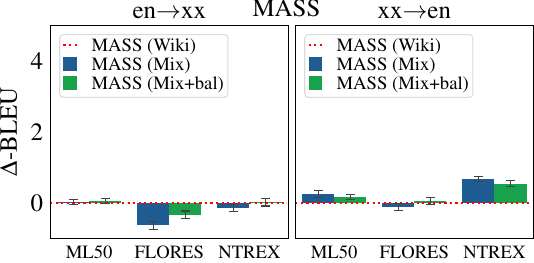}
        \end{subfigure}

        \caption{BLEU differences ($\Delta$-BLEU) of the BT models trained with the mixed-domain split with respect to the single-domain monolingual data (dotted red line). To plot the bars, we use the mean $\Delta$-BLEU and the standard error.}
        \label{fig:delta-bleu-balancing-bt}
    \end{figure*}

    \paragraph{What is the reason for the mixed results?}
    In our experiments, we used the same model/training hyperparameters
    as in previous conflicting studies~\cite{wang-etal-2020-multi,siddhant-etal-2020-leveraging}.
    The only difference lies in our training and test data.
    Those earlier works used 10/15 languages from WMT and news test sets.
    By contrast, the ML50 dataset is more challenging,
    as 1) it has more languages,
    2) contains truly low-resource languages (24/50 have less than
    200K sentences, unlike prior works),
    and, more importantly,  3) it has data from diverse sources (Figure~\ref{fig:ml50-data}).
    High-resource languages contain WMT (news) data,
    whereas other languages have data from different sources, mainly from TED talks.
    Recall that BT is more effective in high-resource pairs
    but yields poor results in non-English non-WMT pairs.
    Considering this, we hypothesize that previous works reported universal gains
    because they considered more favourabe experimental setups, with fewer languages
    and parallel, monolingual, and test data in the same domain.

    \paragraph{How do results change on other test domains?}
    To test this hypothesis, we evaluate models on \textit{uniform} test sets,
    where all languages have data from the same source.
    Figure~\ref{fig:delta-bleu-wiki-flores-ntrex} shows the results on the FLORES (Wikipedia domain)
    and NTREX (news domain) test sets.
    The TICO-19 results follow similar trends and include them in the appendix.

    The results in both FLORES and NTREX reveal a more favorable picture for both methods.
    We see similar trends as in the ML50 test sets, especially in xx$\rightarrow$en,
    but the gains are overall larger.
    This can be explained by the greater domain similarity of the test sets with the monolingual data,
    particularly FLORES, which shows the biggest improvements.
    The switch to the in-domain test sets has a stronger effect on BT,
    especially in en$\rightarrow$xx.
    Notice that in ML50, BT is harmful in en$\rightarrow$xx low-resource with mostly out-of-domain data,
    whereas in NTREX and FLORES, it is consistently helpful.
    MASS also performs much better on the in-domain test data.
    However, we still fail to observe the universal gains reported in some works.
    For instance, in en$\rightarrow$xx, it outperforms the baseline only in low-resource.
    We hypothesize that DAE requires more ideal conditions to be helpful in MMT.
    For instance, \citet{siddhant-etal-2020-leveraging} used much more monolingual relative to the parallel data,
    whereas \citet{wang-etal-2020-multi} used a similar ratio to this work but with parallel, monolingual and test data from the same domain.
    Overall, the performance gap between test sets shows that the domain of the monolingual data is crucial
    and that both methods are sensitive to mismatches with the test domain, particularly BT.

    \subsection{Mixed-domain Monolingual Data}
    \label{sec:result-mixed-mono}

    Previously, we examined single-domain monolingual data, removing confounding factors to isolate domain impact.
    We now turn to a real-world scenario and use multiple sources of monolingual data per language.
    The goal is to evaluate the significance of diversity in monolingual data.
    For each language,
    we hold the size of monolingual data constant (\S\ref{sec:results-wiki-mono}),
    and \textit{only} change the data mixture.
    We include data from News Crawl and CC100 (web domain),
    the only other publicly available data sources with wide enough coverage
    to support most languages in ML50.
    For languages that do not have data from all domains, we use only the available ones.
    We consider two mixed-domain splits:

    \setlist[enumerate]{leftmargin=20pt}
    \begin{enumerate}
    [topsep=3pt,itemsep=4pt,partopsep=0pt, parsep=0pt]
        \item \textit{Unbalanced:} This split emulates naively concatenating all the monolingual data of a given language without considering their relative sizes.
        The ratio between sources is proportional to the size of their uncapped data.
        \item \textit{Balanced:} This split balances the number of sentences from each source using the same temperature-based sampling method applied to the parallel data, with T=5.
    \end{enumerate}

    In Figure~\ref{fig:delta-bleu-balancing-bt},
    each bar shows the average BLEU difference ($\Delta$-BLEU) compared to the single-domain split (Wiki).
    We include results on the TICO-19 and with ChrF scores in the appendix.
    Diversity largely favours BT with a minor impact on MASS.
    This further supports that BT is more sensitive to the domain.
    BT displays a contrast between translation directions.
    Note that 1) both BT and DAE use identical target-side monolingual data,
    and 2) the MMT model has been exposed to a large number of diverse (i.e., many domains)
    English target-side sentences through the ML50 parallel data.
    Thus, we hypothesize that source-side diversity causes the xx$\rightarrow$en gains of BT.

    The highest gains appear in NTREX test sets (up to +4 BLEU),
    as mixed splits incorporate monolingual data from the same domain, i.e., news.
    Interestingly, mixed-domain data proves beneficial for xx$\rightarrow$en in FLORES.
    Closer examination reveals that these gains mainly affect low-resource languages (Table~\ref{tab:mmt-full-main-flores}).
    Although the reason isn't clear,
    we speculate it may be due to reduced cross-domain interference between the parallel and monolingual data.
    The re-balancing of monolingual data has minimal impact,
    though it does slightly enhance or mitigate the drawbacks of using less in-domain data (e.g., FLORES).
    NTREX does not benefit because re-balancing leads to using less news data.

    \begingroup
\setlength{\tabcolsep}{3.7pt} %
\renewcommand{\arraystretch}{1.1} %
\begin{table}
	\small
	\centering
		
	\begin{tabular}{lrrrrrrr}
		\toprule[1.5pt]
		\multirow{2}[3]{*}{Model} &
		\multicolumn{3}{c}{en$\rightarrow$xx} &
		\multicolumn{3}{c}{xx$\rightarrow$en} &
		\multicolumn{1}{c}{\multirow{2}[3]{*}{Mean}}
		\\
		\cmidrule(lr){2-4} \cmidrule(lr){5-7}
		&
		High & Med & Low &
		High & Med & Low &
		\\
				        
		\midrule
		\textit{FLORES}  &               &               &               &               &               &               &               \\
		BART             & 23.6          & 15.2          & 14.3          & 27.8          & 24.4          & 22.0          & 20.8          \\
		MASS             & \textbf{23.8} & \textbf{15.3} & \textbf{14.5} & \textbf{28.4} & \textbf{25.0} & \textbf{23.4} & \textbf{21.3} \\
				
		\midrule
		\textit{NTREX}   &               &               &               &               &               &               &               \\
		BART             & 21.8          & 13.1          & \textbf{13.7} & 25.0          & 23.2          & 21.0          & 19.4          \\
		MASS             & \textbf{21.9} & \textbf{13.3} & \textbf{13.7} & \textbf{25.7} & \textbf{23.8} & \textbf{22.1} & \textbf{19.8} \\
				
		\midrule
		\textit{ML50}    &               &               &               &               &               &               &               \\
		BART             & \textbf{22.1} & \textbf{16.8} & 21.3          & 26.8          & 26.1          & 28.1          & 23.5          \\
		MASS             & \textbf{22.1} & \textbf{16.8} & \textbf{21.5} & \textbf{27.2} & \textbf{26.6} & \textbf{28.8} & \textbf{23.8} \\
				
		\midrule
		\textit{TICO-19} &               &               &               &               &               &               &               \\
		BART             & 31.2          & 14.0          & 15.1          & 31.8          & 26.0          & 24.1          & 23.7          \\
		MASS             & \textbf{31.5} & \textbf{14.4} & \textbf{15.2} & \textbf{32.6} & \textbf{27.2} & \textbf{26.4} & \textbf{24.5} \\
				        
		\bottomrule[1.5pt]
				
	\end{tabular}
	\caption{BLEU scores ($\uparrow$) of BART and MASS trained with the balanced mixed-domain monolingual data.}
	\label{tab:mmt-ssl}
\end{table}
\endgroup

    \subsection{Denoising Autoencoding Objectives}
    \label{sec:ssl}

    Table~\ref{tab:mmt-ssl} compares MASS and BART across all test sets.
    We consider their variants trained with the balanced monolingual data (\S\ref{sec:result-mixed-mono}),
    as they work marginally better (see Appendix \S\ref{sec:app-dae} for more results).
    MASS consistently outperforms BART, with larger gains in xx$\rightarrow$en (up to 2 BLEU).
    However, in xx$\rightarrow$en, their results are comparable.

    Both objectives use similar encoder noising methods but differ in the decoder.
    BART's decoder conditions on the full target prefix, unlike MASS, which excludes unmasked tokens.
    This potentially makes the MASS decoder rely more on its encoder.
    Next, BART computes loss over all tokens, even unmasked ones,
    consequently losing part of the useful signal by teaching the model to copy the input.
    MASS, however, calculates loss only on unmasked tokens, targeting the training signal to denoising.
    In related work, \citet{baziotis-etal-2021-exploring} study NMT pretraining
    using BART variants with different input noising methods, such as word replacement or shuffling,
    and present evidence that input masking biases models towards copying the input.
    We speculate that the performance gap between MASS and BART stems from these decoder-side differences.

    \subsection{Scale}
    \label{sec:scale}

    \begin{figure}[t]
        \begin{subfigure}{1\linewidth}
            \centering
            \includegraphics[width=\linewidth]{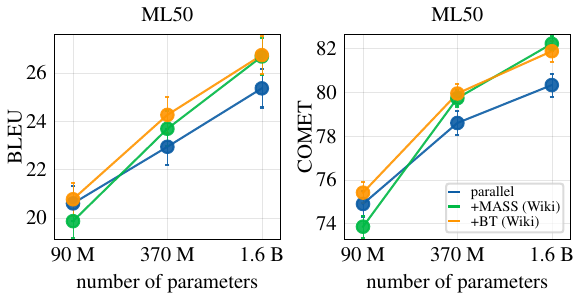}
        \end{subfigure}
        \begin{subfigure}{1\linewidth}
            \centering
            \includegraphics[width=\linewidth]{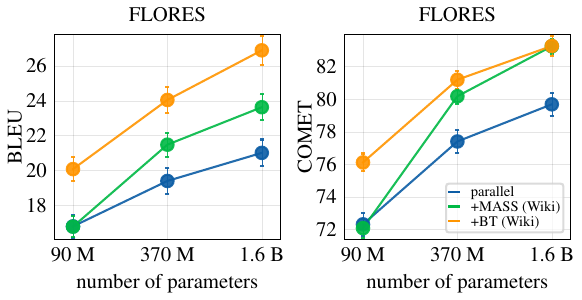}
        \end{subfigure}
        \caption{Mean BLEU and COMET across model scales. The error bars show the standard error of the mean.}
        \label{fig:scale-analysis-methods}
    \end{figure}

    This section examines the role of model scale.
    We hold all other factors constant and test three model sizes that differ by a factor of 4:
    Transformer-Base (90M), Transformer-Big (370M), and Transformer-XL (1.6B).
    To conserve computational resources,
    we consider only one DAE method, MASS, as it outperformed BART in previous experiments.
    We use the (Wiki) single-domain monolingual split
    to test for in-domain (FLORES) and out-of-domain (ML50) effects.
    Figure~\ref{fig:scale-analysis-methods} shows results and includes BLEU and
    COMET\footnote{
        We include COMET here because, whilst in other experiments COMET and BLEU show similar results,
        in this case, we discover a small but noteworthy difference (see Appendix for details; Figures~\ref{fig:scale-line-chrf-all},~\ref{fig:scale-line-comet-all}).}.

    \paragraph{How crucial is model capacity for BT and DAE?}
    All models improve with scale.
    However, small models find monolingual methods less beneficial, especially in ML50 (top),
    which is out-of-domain with respect to the (Wikipedia) monolingual data.
    BT shows negligible gains, while MASS even proves detrimental.
    As scale increases, both MASS and BT become more effective, with MASS benefiting the most.
    Surprisingly, MASS transitions from underperforming the baseline
    to outperforming it and becomes competitive with BT at the 1.6B scale.
    We also discover that according to COMET (and chrF), the effects of scale on MASS are even stronger,
    as it outperforms BT by a small margin.

    In FLORES (bottom), BT and MASS exhibit a similar trend, but are overall more effective,
    since the test and monolingual domains are the same.
    At small scale,  MASS fails to yield any gains, whereas BT is more helpful.
    As scale increases,
    the gains of both methods relative to the baseline also increase.
    However, according to BLEU, the performance gap between MASS and BT remains relatively constant, unlike in ML50,
    whereas according to COMET, MASS achieves again comparable performance to BT.
    This suggests that DAE becomes more competitive with scale and bridges the gap with BT,
    in particular in out-of-domain settings (ML50).

    We speculate that
    learning from monolingual data proves more challenging for smaller models
    because they prioritize learning from parallel data.
    This also explains why BT outperforms DAE at small scales.
    Translating the synthetic parallel data, which is more similar to the supervised MT task, is an easier task compared to denoising.
    As model capacity increases, it ``unblocks'' DAE and progressively enables it to make better use of monolingual data.
    This suggests that there is a cross-task interference that is mitigated by scaling.

    \begin{figure}[t]
        \begin{subfigure}{\linewidth}
            \centering
            \includegraphics[width=0.98\linewidth]{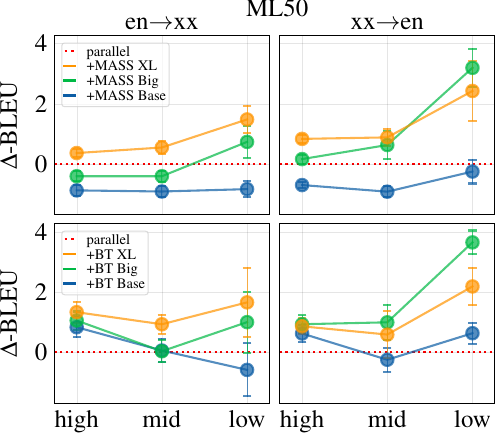}
        \end{subfigure}
        \hfill\par
        \begin{subfigure}{\linewidth}
            \centering
            \includegraphics[width=0.98\linewidth]{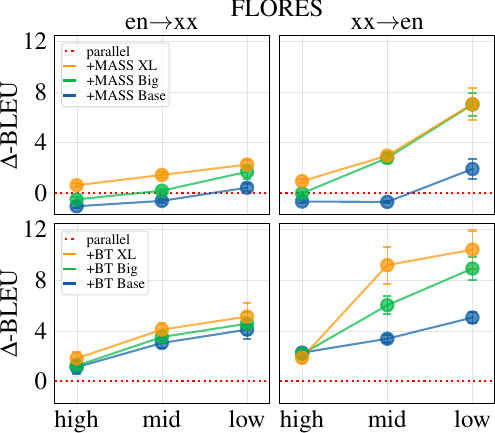}
        \end{subfigure}
        \caption{Average BLEU differences ($\Delta$-BLEU) of each model with respect to the corresponding parallel-only baseline in the same scale (red dotted line).
        The error bars show the standard error of the mean.}
        \label{fig:scale-analysis-methods-detailed}
    \end{figure}

    \paragraph{How direction and resource-level are affected?}

    Next, we investigate the scaling patterns of MASS and BT.
    Figure~\ref{fig:scale-analysis-methods-detailed}
    shows the relative difference between the BLEU score of each model
    and the corresponding parallel-only baseline in the same scale across translation directions.
    Both methods benefit from scale, with low-resource settings gaining the most.
    Notice that for each method, that gap between scales is small in high-resource (up to 2 BLEU) but large in low-resource directions (up to 3 and 5 BLEU in ML50 and FLORES, respectively).
    Scale also generally benefits more xx$\rightarrow$en (right side) compared to en$\rightarrow$xx (left side).
    The plots per test set also have the same y-axis, which enables us to directly compare BT with MASS.
    We discover that the reason MASS (on average) closes the gap with BT (see Figure~\ref{fig:scale-analysis-methods}) as scale grows
    is because of its low-resource performance.
    In particular, in ML50 at the 1.6B scale, the gap becomes negligible,
    and MASS even marginally outperforms BT in low-resource xx$\rightarrow$en (two top-right plots).

    \section{Conclusion}
    \label{sec:conclusion}

    This work presents a systematic analysis of widely used methods that include monolingual data in MMT,
    specifically BT, and two DAE objectives.
    It does not negate findings from prior works but
    rather highlights confounding factors that explain the mixed results found in the literature.
    These factors range from the characteristics of the experimental setup, like the data mixture, to the effective model capacity.
    The main takeaway is that one should not expect gains from DAE or BT in all settings
    but carefully consider all aspects of the system to reach optimal performance.

    We compare models across different data conditions and combinations of monolingual and test data,
    and discover that all methods are very sensitive to domain mismatches.
    BT overall yields the most gains, but it can fail in out-of-domain and low-resource settings.
    As for DAE, we conclude that it can be helpful, particularly in low-resource and xx$\rightarrow$en,
    but the universal gains reported from early works can only be achieved in ideal conditions,
    where the parallel, monolingual, and test data are from the same domain.
    Another key finding is that model capacity can make or break a method.
    Larger models are better able to use monolingual data,
    with gains from both BT and DAE increasing as the model scale grows.
    We also discover a novel connection between domain robustness and model size.
    Scale is more important in out-of-domain settings,
    as all methods yield limited to no gains at small scales.
    In particular, MASS is harmful to MMT with the 90M models,
    but when using 1.6B models, it becomes comparable or even better to BT.

    Based on our findings, we provide some recommendations to practitioners:
    \setlist[itemize]{leftmargin=10pt}
    \begin{itemize} [topsep=3pt,itemsep=4pt,partopsep=0pt, parsep=0pt]
        \item For in-domain settings, prefer BT, as it yields the best results across scales and resource levels.
        \item For out-of-domain settings, the choice depends on model size.
        At small scales, prefer BT but expect small gains.
        At large scales, both methods are more effective, and the gap between them diminishes.
        DAE is a viable and computationally cheaper alternative to BT,
        which needs to backtranslate monolingual data from many languages.
        \item  For MMT+DAE, prefer MASS instead of BART.
        \item Aim to increase the diversity of the monolingual data by mixing different sources and re-balance them to ensure a more even distribution.
        \item If in-domain or diverse monolingual data is not available, consider the trade-offs
        between collecting extra data or scaling up the model.
        If neither is possible, avoid using monolingual data with BT or DAE in en$\rightarrow$xx low-resource directions.
    \end{itemize}

    \section*{Limitations}
    \label{sec:limitations}

    We used only one dataset with roughly 200M sentences and 100 translation directions.
    The dataset is more diverse, with more languages than many prior works, however,
    it is unclear how the results will generalize to datasets with other characteristics,
    such as more languages or more/less typologically diverse languages.
    The same holds for the combinations of monolingual and test data.
    We consider three main sources of publicly available monolingual data
    that also have wide coverage across many languages.
    Using more domains for the monolingual and test data would be better,
    but we could not find other monolingual sources with wide coverage.

    This work focuses only on the English-centric setting
    (i.e., concatenation of English$\rightarrow$XX and XX$\rightarrow$English),
    which is the most commonly studied in MMT and is what the relevant prior works use.
    We considered this setting to make our study directly comparable to those earlier works
    and because it was easier to construct all the different data splits to run both controlled experiments and
    with wide language coverage. However, it is possible that our conclusions do not generalize to other settings,
    such as fully many-to-many MMT or pivot-based MMT.

    This work presents results on three model sizes: 90M, 370M, and 1.6B.
    Our results reveal clear trends emerging across scales,
    but these trends can potentially change in much larger scales depending on the setting.
    One question that is left unanswered is whether DAE would outperform BT if we scaled models to over 1.6B parameters.
    We leave this to future work, as running those experiments would require significantly more resources than we had available.
    On a related note to scale, note that the scale of LLMs is not comparable to MMT models,
    and even models like GPT4 fail to outperform orders of magnitude smaller MMT models like NLLB ( with ``only'' 1.3B)
    in most languages, particularly medium- to low-resource~\cite{zhu2023multilingual}.
    Unlike others, we systematically train models with different methods from-scratch,
    and our larger variant even exceeds the size of models like NLLB.

    Lastly, in this work, we considered the three most widely adopted methods for integrating monolingual data into MMT,
    namely BT and DAE with MASS/BART.
    However, there are other methods, such as those using contrastive losses~\cite{pan-etal-2021-contrastive}.
    We leave these comparisons for future work.

    \section*{Acknowledgments}
    This work was funded by UK Research and Innovation (UKRI) under the UK government’s Horizon Europe funding guarantee
    [grant number 10039436]. The computations described in this research were performed using the Baskerville Tier 2 HPC service (\url{https://www.baskerville.ac.uk/}). Baskerville was funded by the EPSRC and UKRI through the World Class Labs scheme (EP/T022221/1) and the Digital Research Infrastructure programme (EP/W032244/1) and is operated by Advanced Research Computing at the University of Birmingham. We would also like to thank Shruti Bhosale for helpful discussions.

    \bibliography{anthology,custom}
    \bibliographystyle{acl_natbib}

    \clearpage
    \newpage
    \appendix

    \clearpage

    \section{Experimental Setup}
    \label{sec:app-exp-setup}

    \subsection{Training}

    Our baseline is an MMT model trained only on the parallel data (en$\rightarrow$xx and xx$\rightarrow$en).
    For BT, we use the baseline model to generate the synthetic translations using beam search with beam size 4,
    following \citet{nllb2022}.
    For MASS, we use the hyperparameters from \citet{siddhant-etal-2020-leveraging, siddhant2022towards}
    and mask 50\% of input tokens.
    For BART, we use the hyperparameters\footnotemark~from \citet{nllb2022}, that also mask 50\% of input tokens.
    We implement all our models using the fairseq toolkit~\cite{ott2019fairseq}, and for BART we use the original
    implementation in fairseq,
    whereas for MASS develop our own re-implementation.
    \footnotetext{Fairseq arguments: ``--mask 0.5 --mask-random 0.1 \\--mask-length span-poisson --poisson-lambda 3.5``}

    All models use the same Transformer architecture~\cite{vaswani2017attention}
    with shared encoder-decoder embeddings and decoder output projection layers~\cite{press-wolf-2017-using,inan2017tying} as in \citet{nllb2022}.
    We optimize our models with Adam~\cite{kingma2014Adam} with $\beta_1=0.9$, $\beta_2=0.98$, and $\epsilon=10^{-6}$,
    with a learning rate of $0.001$ using a linear warm-up of 8k steps, followed by inverted squared decay.
    We also regularize the models with label smoothing~\cite{szegedy2016rethinking} of 0.1 and weight decay of 0.01.

    We consider three different model sizes:
    1) \textit{Transformer-Base} with 90M parameters configured as in the original paper,
    2) \textit{Transformer-Big} with 370M parameters, similar to the original but with an 8192-sized feed-forward layer
    as in \citet{wang-etal-2020-multi, siddhant-etal-2020-leveraging},
    and 3) \textit{Transformer-XL} (not to be confused with \citet{dai-etal-2019-transformer}),
    with 1.6B parameters, 12 encoder/decoder layers, feed-forward layers of 8192, 2048-sized embeddings, and 32 attention heads.
    We train all models with mixed precision (FP16)
    and use gradient accumulation to reach the desired batch size for each model size.
    Specifically, we train the Transformer-Base on 4 A100 GPUs for 440K steps with an effective batch size of 280K token batches,
    the Transformer-Big on 8 A100 GPUs for 360K steps with 320K token batches,
    and the Transformer-XL on 12 A100 GPUs for 120K steps with 860K token batches.
    We evaluate models every 40K (10k for Transformer-XL) steps and select the checkpoint
    with the best average translation loss (i.e., negative log-likelihood) across all language pairs in the ML50
    validation set.

    \begingroup
    \setlength{\tabcolsep}{5pt} %
    \renewcommand{\arraystretch}{1.1} %
    \begin{table}[t]
        \small
        \centering

        \begin{tabular}{lrrr}
            \toprule[1.5pt]

            & Base          & Big           & XL             \\

            \midrule
            Parameters (Size)    & 90M           & 370M          & 1.6B           \\
            Layers               & 6             & 6             & 12             \\
            Embedding            & 512           & 1024          & 2048           \\
            FeedForward          & 2048          & 8192          & 8192           \\
            Heads                & 8             & 16            & 32             \\
            Effective batch size & 280K          & 320K          & 860K           \\
            Training Steps       & 440K          & 360K          & 120K           \\
            Dropout              & 0.1           & 0.3           & 0.3            \\
            GPU Configuration    & 4$\times$A100 & 8$\times$A100 & 12$\times$A100 \\

            \bottomrule[1.5pt]

        \end{tabular}
        \caption{Hyperparameters used for the Transformer models of various sizes in the study.}
        \label{tab:model-hyperparameters}
    \end{table}
    \endgroup

    \section{Additional Results}
    In the main paper,
    for brevity,
    we discuss results using only BLEU and for selected experiments that highlight our most important findings.
    For completeness, we also re-evaluate the outputs from \textit{all} of our experiments and across all test sets with two additional evaluation metrics, following the recommendations of~\citet{kocmi-etal-2021-ship}:

    \paragraph{chrF:} this is another surface-level (i.e., string-based) metric, like BLEU, but achieves better correlation with human judgment.
    It compares character n-grams that make it better for languages with rich morphology and is also tokenization independent.

    \paragraph{COMET:} this is a neural-based metric that uses a pretrained model to estimate the translation quality.
    Unlike BLEU and chrF, it also takes into account the source sentence.
    However, we point out that it is not clear how reliable (the current version of) COMET is for low-resource languages
    or test data across different domains, as~\citet{kocmi-etal-2021-ship} in their analysis considered only high-resource languages and two test domains (news, discussions).

    We find that overall, the \textit{ranking} of the models is very consistent across metrics.
    We observe only two instances where metrics do not fully agree with each other,
    mainly in en$\rightarrow$xx and low-resource languages
    (see \S\ref{sec:app-mixed-domain}, \S\ref{sec:app-scaling}).
    However, the main findings and patterns discussed in the main body of the paper still hold across metrics.

    \subsection{Main Experiments}
    First, we report the results of the experiments that investigate the role of data.
    This includes the results from all models trained with the single-domain (Wikipedia) and
    mixed-domain (unbalance-vs-balanced) monolingual data in Section~\ref{sec:results-wiki-mono} and Section~\ref{sec:result-mixed-mono}, respectively.
    Recall that ML50 contains parallel data from many different sources, which are mostly out-of-domain data with respect to the Wikipedia domain. The same holds for the ML50 test data.

    We include the full results for all methods across all monolingual splits in
    Table~\ref{tab:mmt-full-main-ml50} (ML50),
    Table~\ref{tab:mmt-full-main-flores} (FLORES),
    Table~\ref{tab:mmt-full-main-ntrex} (NTREX) and Table~\ref{tab:mmt-full-main-tico19} (TICO-19).
    Next, we also include the line charts with the score differences of all models with all metrics in Figure~\ref{fig:delta-scores-all}, which are the counterparts of the Figures~\ref{fig:delta-bleu-wiki},~\ref{fig:delta-bleu-wiki-flores-ntrex} in the main body of the paper.

    \subsubsection{Mixed-Domain Monolingual Data}
    \label{sec:app-mixed-domain}
\begin{figure*}[t!]

        \begin{minipage}[t]{0.48\linewidth}

            \begin{subfigure}{\linewidth}
                \centering
                \includegraphics[width=\linewidth]{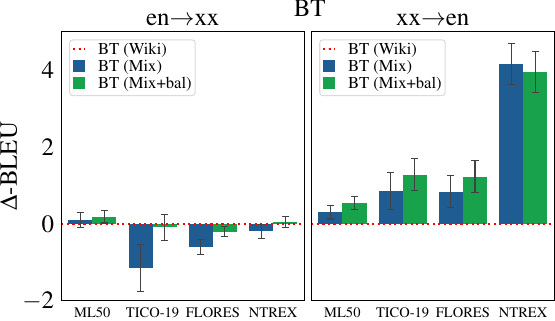}
            \end{subfigure}
            \hfill\par\vspace*{12pt}

            \begin{subfigure}{\linewidth}
                \centering
                \includegraphics[width=\linewidth]{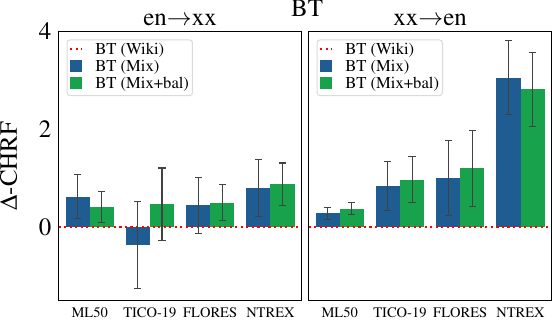}
            \end{subfigure}
            \hfill\par\vspace*{12pt}

            \begin{subfigure}{\linewidth}
                \centering
                \includegraphics[width=\linewidth]{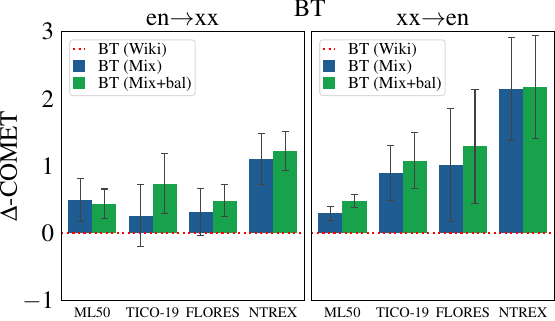}
            \end{subfigure}

            \hfill\par\vspace*{4pt}

            \caption{Score differences ($\Delta$-X) of the \textbf{BT models} trained with the mixed-domain split with respect to the single-domain monolingual data (dotted red line). The top plot shows the $\Delta$-BLEU scores, whereas the bottom shows the $\Delta$-ChrF scores.
            To plot the bars, we use the mean $\Delta$-X and the standard error.}
            \label{fig:appendix-mix-analysis-bleu}
        \end{minipage}
        \hfill
        \begin{minipage}[t]{0.48\linewidth}

            \begin{subfigure}{\linewidth}
                \centering
                \includegraphics[width=\linewidth]{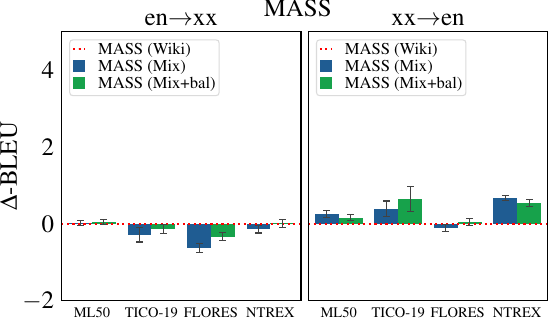}
            \end{subfigure}
            \hfill\par\vspace*{10pt}

            \begin{subfigure}{\linewidth}
                \centering
                \includegraphics[width=\linewidth]{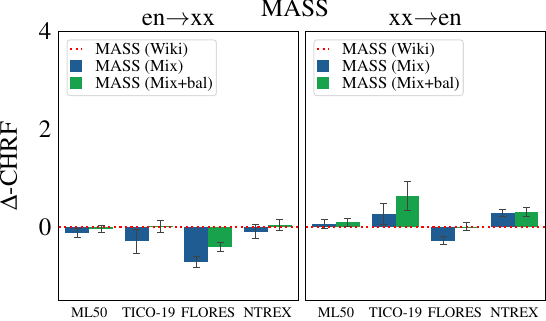}
            \end{subfigure}
            \hfill\par\vspace*{10pt}

            \begin{subfigure}{\linewidth}
                \centering
                \includegraphics[width=\linewidth]{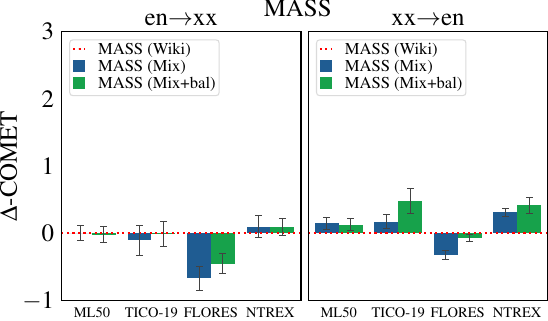}
            \end{subfigure}

            \hfill\par\vspace*{4pt}

            \caption{Score differences ($\Delta$-X) of the \textbf{MASS (DAE) models} trained with the mixed-domain split with respect to the single-domain monolingual data (dotted red line). The top plot shows the $\Delta$-BLEU scores, whereas the bottom shows the $\Delta$-ChrF scores.
            To plot the bars, we use the mean $\Delta$-X and the standard error.}

            \label{fig:appendix-mix-analysis-chrf}
        \end{minipage}

    \end{figure*}

    Besides the table view of the results, which do include the scores per monolingual split,
    here we also report the corresponding bar plots, similar to those in Section~\ref{sec:result-mixed-mono},
    with all methods, test sets, and metrics.
    This is one of the few cases where we discover a small discrepancy between metrics.
    Specifically, we see that the ChrF and COMET results suggest that using mixed-domain monolingual data
    is even \textit{more helpful} for BT, than what the BLEU scores suggest.
    In particular, Figure~\ref{fig:appendix-mix-analysis-bleu} shows gains in BLEU (top) only in the xx$\rightarrow$en direction,
    whereas the ChrF (middle row) and COMET (bottom row) scores reveal consistent improvements even in the en$\rightarrow$xx direction.
    We also see that further re-balancing (green bar) the monolingual data yields small gains in most settings.
    Besides these differences, the overall trends are the same across metrics (i.e., BT is more sensitive to diversity than MASS, with larger gains in xx$\rightarrow$en).

    \begin{table*}[bt]
        \centering
        \begin{minipage}{0.32\textwidth}
            \centering
            
\setlength{\tabcolsep}{1.5pt} %
\renewcommand{\arraystretch}{0.75} %
\scriptsize
\begin{tabular}{lrrrrrrr}
\toprule[1.5pt]
\multirow{2}[3]{*}{Model} &
\multicolumn{3}{c}{en$\rightarrow$xx} &
\multicolumn{3}{c}{xx$\rightarrow$en} &
\multicolumn{1}{c}{\multirow{2}[3]{*}{Mean}}
\\
\cmidrule(lr){2-4} \cmidrule(lr){5-7}
                        &                     High &                      Med &           Low &                     High &           Med & \multicolumn{2}{l}{Low} \\
\midrule
               parallel &                     22.5 &                     17.3 &          20.6 &                     26.9 &          25.9 &          25.3 &          22.9 \\
\addlinespace
\textit{Wiki} & & & & & & & \\
   +BART  & \cellcolor{red!15}{22.0} & \cellcolor{red!15}{16.9} &          21.3 &                     27.0 &          26.6 &          27.9 &          23.6 \\
   +MASS  & \cellcolor{red!15}{22.1} & \cellcolor{red!15}{16.9} &          21.3 &                     27.1 &          26.5 &          28.5 &          23.7 \\
     +BT  &                     23.6 &                     17.3 & \textbf{21.6} &                     27.8 &          26.9 &          28.9 &          24.3 \\
\addlinespace
\textit{Mix} & & & & & & & \\
    +BART  & \cellcolor{red!15}{21.8} & \cellcolor{red!15}{16.7} &          21.4 &                     27.1 &          26.3 &          28.4 &          23.6 \\
    +MASS  & \cellcolor{red!15}{22.0} & \cellcolor{red!15}{16.8} &          21.5 &                     27.4 &          26.5 &          28.9 &          23.8 \\
      +BT  &                     24.0 &            \textbf{17.5} &          21.3 &                     28.3 &          26.9 &          29.4 &          24.5 \\
\addlinespace
\textit{Mix+bal} & & & & & & & \\
+BART  & \cellcolor{red!15}{22.1} & \cellcolor{red!15}{16.8} &          21.3 & \cellcolor{red!15}{26.8} &          26.1 &          28.1 &          23.5 \\
+MASS  & \cellcolor{red!15}{22.1} & \cellcolor{red!15}{16.8} &          21.5 &                     27.2 &          26.6 &          28.8 &          23.8 \\
  +BT  &            \textbf{24.1} &            \textbf{17.5} &          21.4 &            \textbf{28.5} & \textbf{27.2} & \textbf{29.6} & \textbf{24.6} \\
\bottomrule
\end{tabular}

            \vspace{-4pt}
            \subcaption{BLEU scores ($\uparrow$)}
            \label{tab:mmt-full-main-ml50-bleu}
        \end{minipage}\hfill
        \begin{minipage}{0.32\textwidth}
            \centering
            
\setlength{\tabcolsep}{1.5pt} %
\renewcommand{\arraystretch}{0.75} %
\scriptsize
\begin{tabular}{lrrrrrrr}
\toprule[1.5pt]
\multirow{2}[3]{*}{Model} &
\multicolumn{3}{c}{en$\rightarrow$xx} &
\multicolumn{3}{c}{xx$\rightarrow$en} &
\multicolumn{1}{c}{\multirow{2}[3]{*}{Mean}}
\\
\cmidrule(lr){2-4} \cmidrule(lr){5-7}
                        &                     High &                      Med &           Low &                     High &           Med & \multicolumn{2}{l}{Low} \\
\midrule
               parallel &                     48.6 &                     45.0 &          46.3 &                     55.3 &          49.3 &          47.0 &          48.3 \\
\addlinespace
\textit{Wiki} & & & & & & & \\
   +BART  & \cellcolor{red!15}{47.9} & \cellcolor{red!15}{44.6} &          47.4 & \cellcolor{red!15}{55.1} &          50.5 &          50.4 &          49.1 \\
   +MASS  & \cellcolor{red!15}{48.1} & \cellcolor{red!15}{44.5} &          47.5 & \cellcolor{red!15}{55.2} &          50.6 &          50.9 &          49.3 \\
     +BT  &                     49.7 &                     45.4 &          47.2 &                     56.4 &          51.6 &          51.7 &          50.1 \\
\addlinespace
\textit{Mix} & & & & & & & \\
    +BART  & \cellcolor{red!15}{47.7} & \cellcolor{red!15}{44.0} &          47.4 &                     55.3 &          50.4 &          50.8 &          49.1 \\
    +MASS  & \cellcolor{red!15}{47.9} & \cellcolor{red!15}{44.4} &          47.5 &                     55.3 &          50.4 &          51.2 &          49.3 \\
      +BT  &                     49.8 &            \textbf{46.1} & \textbf{48.0} &                     56.3 &          51.6 &          52.4 & \textbf{50.5} \\
\addlinespace
\textit{Mix+bal} & & & & & & & \\
+BART  & \cellcolor{red!15}{47.9} & \cellcolor{red!15}{44.1} &          47.4 & \cellcolor{red!15}{55.1} &          50.4 &          50.6 &          49.1 \\
+MASS  & \cellcolor{red!15}{48.0} & \cellcolor{red!15}{44.4} &          47.6 & \cellcolor{red!15}{55.2} &          50.6 &          51.2 &          49.3 \\
  +BT  &            \textbf{50.0} &                     45.7 &          47.8 &            \textbf{56.5} & \textbf{51.7} & \textbf{52.5} & \textbf{50.5} \\
\bottomrule
\end{tabular}

            \vspace{-4pt}
            \subcaption{chrF scores ($\uparrow$)}
            \label{tab:mmt-full-main-ml50-chrf}
        \end{minipage}\hfill
        \begin{minipage}{0.32\textwidth}
            \centering
            
\setlength{\tabcolsep}{1.5pt} %
\renewcommand{\arraystretch}{0.75} %
\scriptsize
\begin{tabular}{lrrrrrrr}
\toprule[1.5pt]
\multirow{2}[3]{*}{Model} &
\multicolumn{3}{c}{en$\rightarrow$xx} &
\multicolumn{3}{c}{xx$\rightarrow$en} &
\multicolumn{1}{c}{\multirow{2}[3]{*}{Mean}}
\\
\cmidrule(lr){2-4} \cmidrule(lr){5-7}
                        &                     High &                      Med &           Low &          High &           Med & \multicolumn{2}{l}{Low} \\
\midrule
               parallel &                     80.9 &                     80.5 &          77.0 &          80.8 &          78.0 &          75.6 &          78.6 \\
\addlinespace
\textit{Wiki} & & & & & & & \\
   +BART  & \cellcolor{red!15}{80.4} & \cellcolor{red!15}{80.2} &          78.4 &          80.9 &          79.1 &          78.9 &          79.5 \\
   +MASS  & \cellcolor{red!15}{80.7} & \cellcolor{red!15}{79.9} &          78.6 &          81.0 &          79.4 &          79.4 &          79.7 \\
     +BT  &                     81.8 &                     80.9 &          78.3 &          81.3 &          79.4 &          78.9 &          80.0 \\
\addlinespace
\textit{Mix} & & & & & & & \\
    +BART  & \cellcolor{red!15}{80.0} & \cellcolor{red!15}{79.2} &          78.7 &          80.9 &          79.0 &          79.3 &          79.4 \\
    +MASS  & \cellcolor{red!15}{80.5} & \cellcolor{red!15}{79.9} & \textbf{78.8} &          81.2 &          79.3 & \textbf{79.8} &          79.8 \\
      +BT  &                     82.2 &            \textbf{81.6} &          78.7 &          81.4 &          79.4 &          79.6 &          80.3 \\
\addlinespace
\textit{Mix+bal} & & & & & & & \\
+BART  & \cellcolor{red!15}{80.2} & \cellcolor{red!15}{79.5} &          78.3 &          80.8 &          78.9 &          79.0 &          79.4 \\
+MASS  & \cellcolor{red!15}{80.6} & \cellcolor{red!15}{79.7} & \textbf{78.8} &          81.2 &          79.3 & \textbf{79.8} &          79.8 \\
  +BT  &            \textbf{82.3} &                     81.2 & \textbf{78.8} & \textbf{81.6} & \textbf{79.6} & \textbf{79.8} & \textbf{80.4} \\
\bottomrule
\end{tabular}

            \vspace{-4pt}
            \subcaption{COMET scores ($\uparrow$)}
            \label{tab:mmt-full-main-ml50-comet}
        \end{minipage}
        \caption{Results of the Transformer-Big models evaluated on the \textbf{ML50} (mixed-domain) test set and grouped by the monolingual split that has been used for training BT and DAE.}
        \label{tab:mmt-full-main-ml50}
    \end{table*}

    \begin{table*}[bt]
        \centering
        \begin{minipage}{0.32\textwidth}
            \centering
            
\setlength{\tabcolsep}{1.5pt} %
\renewcommand{\arraystretch}{0.75} %
\scriptsize
\begin{tabular}{lrrrrrrr}
\toprule[1.5pt]
\multirow{2}[3]{*}{Model} &
\multicolumn{3}{c}{en$\rightarrow$xx} &
\multicolumn{3}{c}{xx$\rightarrow$en} &
\multicolumn{1}{c}{\multirow{2}[3]{*}{Mean}}
\\
\cmidrule(lr){2-4} \cmidrule(lr){5-7}
                        &                     High &                      Med &           Low &                     High &           Med & \multicolumn{2}{l}{Low} \\
\midrule
               parallel &                     24.8 &                     15.3 &          13.2 &                     28.6 &          22.4 &          16.0 &          19.4 \\
\addlinespace
\textit{Wiki} & & & & & & & \\
   +BART  & \cellcolor{red!15}{24.0} &                     15.6 &          14.7 & \cellcolor{red!15}{28.3} &          24.9 &          22.5 &          21.2 \\
   +MASS  & \cellcolor{red!15}{24.3} &                     15.5 &          14.9 &                     28.6 &          25.2 &          23.0 &          21.5 \\
     +BT  &            \textbf{26.0} &            \textbf{18.8} &          17.7 &                     30.8 &          28.4 &          24.9 &          24.1 \\
\addlinespace
\textit{Mix} & & & & & & & \\
    +BART  & \cellcolor{red!15}{23.4} & \cellcolor{red!15}{15.0} &          14.2 & \cellcolor{red!15}{27.9} &          24.7 &          22.6 &          20.9 \\
    +MASS  & \cellcolor{red!15}{23.5} & \cellcolor{red!15}{15.2} &          14.1 & \cellcolor{red!15}{28.5} &          24.8 &          23.0 &          21.1 \\
      +BT  &                     25.6 &                     17.6 &          17.6 &                     30.8 &          28.6 &          26.9 &          24.2 \\
\addlinespace
\textit{Mix+bal} & & & & & & & \\
+BART  & \cellcolor{red!15}{23.6} & \cellcolor{red!15}{15.2} &          14.3 & \cellcolor{red!15}{27.8} &          24.4 &          22.0 &          20.8 \\
+MASS  & \cellcolor{red!15}{23.8} &                     15.3 &          14.5 & \cellcolor{red!15}{28.4} &          25.0 &          23.4 &          21.3 \\
  +BT  &                     25.5 &                     18.3 & \textbf{18.0} &            \textbf{31.1} & \textbf{29.1} & \textbf{27.2} & \textbf{24.6} \\
\bottomrule
\end{tabular}

            \vspace{-4pt}
            \subcaption{BLEU scores ($\uparrow$)}
            \label{tab:mmt-full-main-flores-bleu}
        \end{minipage}\hfill
        \begin{minipage}{0.32\textwidth}
            \centering
            
\setlength{\tabcolsep}{1.5pt} %
\renewcommand{\arraystretch}{0.75} %
\scriptsize
\begin{tabular}{lrrrrrrr}
\toprule[1.5pt]
\multirow{2}[3]{*}{Model} &
\multicolumn{3}{c}{en$\rightarrow$xx} &
\multicolumn{3}{c}{xx$\rightarrow$en} &
\multicolumn{1}{c}{\multirow{2}[3]{*}{Mean}}
\\
\cmidrule(lr){2-4} \cmidrule(lr){5-7}
                        &                     High &                      Med &           Low &                     High &           Med & \multicolumn{2}{l}{Low} \\
\midrule
               parallel &                     50.6 &                     46.1 &          45.0 &                     57.1 &          50.4 &          42.6 &          48.2 \\
\addlinespace
\textit{Wiki} & & & & & & & \\
   +BART  & \cellcolor{red!15}{49.9} &                     46.2 &          47.1 & \cellcolor{red!15}{56.9} &          53.1 &          50.6 &          50.4 \\
   +MASS  & \cellcolor{red!15}{50.2} &                     46.1 &          47.3 & \cellcolor{red!15}{57.0} &          53.4 &          51.3 &          50.6 \\
     +BT  &            \textbf{52.1} &            \textbf{49.3} &          47.6 &                     59.2 &          57.1 &          52.2 &          52.7 \\
\addlinespace
\textit{Mix} & & & & & & & \\
    +BART  & \cellcolor{red!15}{49.5} & \cellcolor{red!15}{45.1} &          46.4 & \cellcolor{red!15}{56.5} &          52.9 &          50.6 &          49.9 \\
    +MASS  & \cellcolor{red!15}{49.4} & \cellcolor{red!15}{45.7} &          46.4 & \cellcolor{red!15}{56.7} &          53.0 &          51.2 &          50.1 \\
      +BT  &                     51.6 &                     49.2 & \textbf{49.3} &                     59.1 &          57.0 &          55.1 &          53.4 \\
\addlinespace
\textit{Mix+bal} & & & & & & & \\
+BART  & \cellcolor{red!15}{49.4} & \cellcolor{red!15}{45.3} &          46.5 & \cellcolor{red!15}{56.4} &          52.9 &          50.4 &          49.9 \\
+MASS  & \cellcolor{red!15}{49.7} & \cellcolor{red!15}{45.8} &          46.8 & \cellcolor{red!15}{56.8} &          53.3 &          51.6 &          50.4 \\
  +BT  &                     51.7 &                     49.2 & \textbf{49.3} &            \textbf{59.3} & \textbf{57.3} & \textbf{55.2} & \textbf{53.5} \\
\bottomrule
\end{tabular}

            \vspace{-4pt}
            \subcaption{chrF scores ($\uparrow$)}
            \label{tab:mmt-full-main-flores-chrf}
        \end{minipage}\hfill
        \begin{minipage}{0.32\textwidth}
            \centering
            
\setlength{\tabcolsep}{1.5pt} %
\renewcommand{\arraystretch}{0.75} %
\scriptsize
\begin{tabular}{lrrrrrrr}
\toprule[1.5pt]
\multirow{2}[3]{*}{Model} &
\multicolumn{3}{c}{en$\rightarrow$xx} &
\multicolumn{3}{c}{xx$\rightarrow$en} &
\multicolumn{1}{c}{\multirow{2}[3]{*}{Mean}}
\\
\cmidrule(lr){2-4} \cmidrule(lr){5-7}
                        &                     High &                      Med &           Low &                     High &           Med & \multicolumn{2}{l}{Low} \\
\midrule
               parallel &                     83.0 &                     80.5 &          71.4 &                     84.0 &          79.3 &          69.9 &          77.4 \\
\addlinespace
\textit{Wiki} & & & & & & & \\
   +BART  & \cellcolor{red!15}{82.5} &                     80.5 &          74.4 & \cellcolor{red!15}{83.9} &          81.5 &          78.9 &          80.0 \\
   +MASS  & \cellcolor{red!15}{82.8} & \cellcolor{red!15}{80.3} &          74.6 & \cellcolor{red!15}{83.9} &          81.7 &          79.6 &          80.2 \\
     +BT  &                     84.1 &                     82.7 &          77.5 &            \textbf{84.8} &          82.9 &          77.2 &          81.2 \\
\addlinespace
\textit{Mix} & & & & & & & \\
    +BART  & \cellcolor{red!15}{81.8} & \cellcolor{red!15}{79.1} &          73.5 & \cellcolor{red!15}{83.4} &          81.1 &          78.6 &          79.3 \\
    +MASS  & \cellcolor{red!15}{82.0} & \cellcolor{red!15}{80.0} &          73.7 & \cellcolor{red!15}{83.7} &          81.3 &          79.3 &          79.7 \\
      +BT  &                     84.1 &            \textbf{83.0} &          78.1 &                     84.6 &          82.7 &          80.1 &          81.9 \\
\addlinespace
\textit{Mix+bal} & & & & & & & \\
+BART  & \cellcolor{red!15}{81.9} & \cellcolor{red!15}{79.5} &          73.4 & \cellcolor{red!15}{83.3} &          81.0 &          78.5 &          79.3 \\
+MASS  & \cellcolor{red!15}{82.3} & \cellcolor{red!15}{80.0} &          74.0 & \cellcolor{red!15}{83.7} &          81.6 &          79.7 &          79.9 \\
  +BT  &            \textbf{84.2} &                     82.9 & \textbf{78.6} &            \textbf{84.8} & \textbf{83.0} & \textbf{80.4} & \textbf{82.1} \\
\bottomrule
\end{tabular}

            \vspace{-4pt}
            \subcaption{COMET scores ($\uparrow$)}
            \label{tab:mmt-full-main-flores-comet}
        \end{minipage}
        \caption{Results of the Transformer-Big models on the \textbf{FLORES} (Wikipedia) test set and grouped by the monolingual split that has been used for training BT and DAE. Cells in \colorbox{red!15}{red} indicate worse scores than the baseline.}
        \label{tab:mmt-full-main-flores}
    \end{table*}

    \begin{table*}[bt]
        \centering
        \begin{minipage}{0.32\textwidth}
            \centering
            
\setlength{\tabcolsep}{1.5pt} %
\renewcommand{\arraystretch}{0.75} %
\scriptsize
\begin{tabular}{lrrrrrrr}
\toprule[1.5pt]
\multirow{2}[3]{*}{Model} &
\multicolumn{3}{c}{en$\rightarrow$xx} &
\multicolumn{3}{c}{xx$\rightarrow$en} &
\multicolumn{1}{c}{\multirow{2}[3]{*}{Mean}}
\\
\cmidrule(lr){2-4} \cmidrule(lr){5-7}
                        &                     High &                      Med &           Low &                     High &           Med & \multicolumn{2}{l}{Low} \\
\midrule
               parallel &                     22.4 &                     13.2 &          12.4 &                     25.1 &          21.1 &          15.1 &          17.8 \\
\addlinespace
\textit{Wiki} & & & & & & & \\
   +BART  & \cellcolor{red!15}{21.9} &                     13.2 &          13.4 &                     25.5 &          23.3 &          20.8 &          19.4 \\
   +MASS  & \cellcolor{red!15}{22.1} &                     13.2 &          13.8 &                     25.5 &          23.3 &          21.2 &          19.5 \\
     +BT  &            \textbf{23.3} &            \textbf{15.5} &          16.0 &                     27.4 &          25.1 &          21.8 &          21.2 \\
\addlinespace
\textit{Mix} & & & & & & & \\
    +BART  & \cellcolor{red!15}{21.6} & \cellcolor{red!15}{13.0} &          13.6 &                     25.4 &          23.4 &          21.5 &          19.5 \\
    +MASS  & \cellcolor{red!15}{21.7} & \cellcolor{red!15}{13.1} &          13.7 &                     26.0 &          23.9 &          22.1 &          19.8 \\
      +BT  &                     22.8 &                     14.9 &          16.4 &            \textbf{30.9} & \textbf{28.6} & \textbf{27.1} & \textbf{23.2} \\
\addlinespace
\textit{Mix+bal} & & & & & & & \\
+BART  & \cellcolor{red!15}{21.8} & \cellcolor{red!15}{13.1} &          13.7 & \cellcolor{red!15}{25.0} &          23.2 &          21.0 &          19.4 \\
+MASS  & \cellcolor{red!15}{21.9} &                     13.3 &          13.7 &                     25.7 &          23.8 &          22.1 &          19.8 \\
  +BT  &                     22.9 &                     15.4 & \textbf{16.6} &                     30.4 &          28.5 &          27.0 & \textbf{23.2} \\
\bottomrule
\end{tabular}

            \vspace{-4pt}
            \subcaption{BLEU scores ($\uparrow$)}
            \label{tab:mmt-full-main-ntrex-bleu}
        \end{minipage}\hfill
        \begin{minipage}{0.32\textwidth}
            \centering
            
\setlength{\tabcolsep}{1.5pt} %
\renewcommand{\arraystretch}{0.75} %
\scriptsize
\begin{tabular}{lrrrrrrr}
\toprule[1.5pt]
\multirow{2}[3]{*}{Model} &
\multicolumn{3}{c}{en$\rightarrow$xx} &
\multicolumn{3}{c}{xx$\rightarrow$en} &
\multicolumn{1}{c}{\multirow{2}[3]{*}{Mean}}
\\
\cmidrule(lr){2-4} \cmidrule(lr){5-7}
                        &                     High &                      Med &           Low &                     High &           Med & \multicolumn{2}{l}{Low} \\
\midrule
               parallel &                     48.3 &                     43.3 &          42.2 &                     54.5 &          49.3 &          40.8 &          46.0 \\
\addlinespace
\textit{Wiki} & & & & & & & \\
   +BART  & \cellcolor{red!15}{47.6} & \cellcolor{red!15}{43.2} &          44.0 &                     54.5 &          51.7 &          47.9 &          47.9 \\
   +MASS  & \cellcolor{red!15}{47.9} & \cellcolor{red!15}{43.0} &          44.2 &                     54.6 &          51.9 &          48.5 &          48.1 \\
     +BT  &            \textbf{49.2} &                     45.7 &          44.2 &                     56.7 &          54.6 &          48.3 &          49.5 \\
\addlinespace
\textit{Mix} & & & & & & & \\
    +BART  & \cellcolor{red!15}{47.4} & \cellcolor{red!15}{42.8} &          43.9 &                     54.5 &          51.7 &          48.4 &          47.9 \\
    +MASS  & \cellcolor{red!15}{47.5} & \cellcolor{red!15}{43.2} &          44.1 &                     54.7 &          52.0 &          49.1 &          48.2 \\
      +BT  &                     48.8 &                     46.2 & \textbf{46.2} &            \textbf{58.7} & \textbf{56.6} & \textbf{53.1} & \textbf{51.4} \\
\addlinespace
\textit{Mix+bal} & & & & & & & \\
+BART  & \cellcolor{red!15}{47.4} & \cellcolor{red!15}{42.9} &          44.1 & \cellcolor{red!15}{54.4} &          51.8 &          48.2 &          47.9 \\
+MASS  & \cellcolor{red!15}{47.6} &                     43.3 &          44.3 &                     54.6 &          52.1 &          49.3 &          48.3 \\
  +BT  &                     49.1 &            \textbf{46.3} & \textbf{46.2} &                     58.4 &          56.5 &          52.8 & \textbf{51.4} \\
\bottomrule
\end{tabular}

            \vspace{-4pt}
            \subcaption{chrF scores ($\uparrow$)}
            \label{tab:mmt-full-main-ntrex-chrf}
        \end{minipage}\hfill
        \begin{minipage}{0.32\textwidth}
            \centering
            
\setlength{\tabcolsep}{1.5pt} %
\renewcommand{\arraystretch}{0.75} %
\scriptsize
\begin{tabular}{lrrrrrrr}
\toprule[1.5pt]
\multirow{2}[3]{*}{Model} &
\multicolumn{3}{c}{en$\rightarrow$xx} &
\multicolumn{3}{c}{xx$\rightarrow$en} &
\multicolumn{1}{c}{\multirow{2}[3]{*}{Mean}}
\\
\cmidrule(lr){2-4} \cmidrule(lr){5-7}
                        &                     High &                      Med &           Low &                     High &           Med & \multicolumn{2}{l}{Low} \\
\midrule
               parallel &                     79.0 &                     76.9 &          69.4 &                     82.1 &          78.7 &          67.6 &          75.2 \\
\addlinespace
\textit{Wiki} & & & & & & & \\
   +BART  & \cellcolor{red!15}{78.3} & \cellcolor{red!15}{76.7} &          72.0 &                     82.1 &          80.6 &          75.5 &          77.3 \\
   +MASS  & \cellcolor{red!15}{78.8} & \cellcolor{red!15}{76.4} &          72.3 &                     82.3 &          80.8 &          76.3 &          77.6 \\
     +BT  &                     79.7 &                     78.7 &          74.4 &                     83.0 &          81.6 &          73.4 &          78.2 \\
\addlinespace
\textit{Mix} & & & & & & & \\
    +BART  & \cellcolor{red!15}{78.0} & \cellcolor{red!15}{76.0} &          72.0 & \cellcolor{red!15}{81.9} &          80.5 &          76.2 &          77.2 \\
    +MASS  & \cellcolor{red!15}{78.5} &                     76.9 &          72.3 &                     82.3 &          81.0 &          76.9 &          77.8 \\
      +BT  &                     80.1 &                     79.6 &          76.2 &            \textbf{83.8} &          82.6 & \textbf{77.6} &          79.8 \\
\addlinespace
\textit{Mix+bal} & & & & & & & \\
+BART  & \cellcolor{red!15}{78.2} & \cellcolor{red!15}{76.3} &          71.9 & \cellcolor{red!15}{81.8} &          80.5 &          75.9 &          77.2 \\
+MASS  & \cellcolor{red!15}{78.6} & \cellcolor{red!15}{76.8} &          72.3 &                     82.2 &          81.0 &          77.2 &          77.8 \\
  +BT  &            \textbf{80.2} &            \textbf{79.7} & \textbf{76.3} &            \textbf{83.8} & \textbf{82.7} & \textbf{77.6} & \textbf{79.9} \\
\bottomrule
\end{tabular}

            \vspace{-4pt}
            \subcaption{COMET scores ($\uparrow$)}
            \label{tab:mmt-full-main-ntrex-comet}
        \end{minipage}
        \caption{Results of the Transformer-Big models on the \textbf{NTREX} (News) test set and grouped by the monolingual split that has been used for training BT and DAE. Cells in \colorbox{red!15}{red} indicate worse scores than the baseline.}
        \label{tab:mmt-full-main-ntrex}
    \end{table*}

    \begin{table*}[bt]
        \centering
        \begin{minipage}{0.32\textwidth}
            \centering
            
\setlength{\tabcolsep}{1.5pt} %
\renewcommand{\arraystretch}{0.75} %
\scriptsize
\begin{tabular}{lrrrrrrr}
\toprule[1.5pt]
\multirow{2}[3]{*}{Model} &
\multicolumn{3}{c}{en$\rightarrow$xx} &
\multicolumn{3}{c}{xx$\rightarrow$en} &
\multicolumn{1}{c}{\multirow{2}[3]{*}{Mean}}
\\
\cmidrule(lr){2-4} \cmidrule(lr){5-7}
                        &                     High &                      Med &           Low &                     High &           Med & \multicolumn{2}{l}{Low} \\
\midrule
               parallel &                     32.3 &                     14.3 &          14.4 &                     32.4 &          24.2 &          17.4 &          22.3 \\
\addlinespace
\textit{Wiki} & & & & & & & \\
   +BART  & \cellcolor{red!15}{31.9} &                     14.9 &          15.1 &                     32.9 &          26.3 &          24.2 &          24.2 \\
   +MASS  & \cellcolor{red!15}{31.9} & \cellcolor{red!15}{14.0} &          15.4 &                     32.9 &          27.0 &          24.6 &          24.3 \\
     +BT  &            \textbf{34.5} &            \textbf{18.4} &          19.8 &                     36.8 &          32.2 &          28.7 &          28.3 \\
\addlinespace
\textit{Mix} & & & & & & & \\
    +BART  & \cellcolor{red!15}{30.5} & \cellcolor{red!15}{13.9} &          14.9 &                     32.5 &          26.6 &          24.2 &          23.7 \\
    +MASS  & \cellcolor{red!15}{31.1} &                     14.3 &          15.2 &                     33.0 &          26.9 &          25.6 &          24.3 \\
      +BT  &                     33.2 &                     16.5 &          19.2 &                     36.9 &          32.5 &          30.6 &          28.2 \\
\addlinespace
\textit{Mix+bal} & & & & & & & \\
+BART  & \cellcolor{red!15}{31.2} & \cellcolor{red!15}{14.0} &          15.1 & \cellcolor{red!15}{31.8} &          26.0 &          24.1 &          23.7 \\
+MASS  & \cellcolor{red!15}{31.5} &                     14.4 &          15.2 &                     32.6 &          27.2 &          26.4 &          24.5 \\
  +BT  &                     34.3 &                     17.7 & \textbf{20.2} &            \textbf{37.4} & \textbf{33.0} & \textbf{30.9} & \textbf{28.9} \\
\bottomrule
\end{tabular}

            \vspace{-4pt}
            \subcaption{BLEU scores ($\uparrow$)}
            \label{tab:mmt-full-main-tico19-bleu}
        \end{minipage}\hfill
        \begin{minipage}{0.32\textwidth}
            \centering
            
\setlength{\tabcolsep}{1.5pt} %
\renewcommand{\arraystretch}{0.75} %
\scriptsize
\begin{tabular}{lrrrrrrr}
\toprule[1.5pt]
\multirow{2}[3]{*}{Model} &
\multicolumn{3}{c}{en$\rightarrow$xx} &
\multicolumn{3}{c}{xx$\rightarrow$en} &
\multicolumn{1}{c}{\multirow{2}[3]{*}{Mean}}
\\
\cmidrule(lr){2-4} \cmidrule(lr){5-7}
                        &                     High &                      Med &           Low &                     High &           Med & \multicolumn{2}{l}{Low} \\
\midrule
               parallel &                     53.3 &                     45.5 &          46.8 &                     61.0 &          52.4 &          45.4 &          50.6 \\
\addlinespace
\textit{Wiki} & & & & & & & \\
   +BART  & \cellcolor{red!15}{52.8} &                     46.2 &          47.9 &                     61.0 &          54.6 &          53.3 &          52.6 \\
   +MASS  & \cellcolor{red!15}{52.8} & \cellcolor{red!15}{44.8} &          48.0 &                     61.0 &          55.2 &          53.3 &          52.5 \\
     +BT  &            \textbf{55.4} &            \textbf{49.7} &          48.6 &                     64.2 &          60.7 &          57.0 &          55.8 \\
\addlinespace
\textit{Mix} & & & & & & & \\
    +BART  & \cellcolor{red!15}{51.7} & \cellcolor{red!15}{44.5} &          47.6 & \cellcolor{red!15}{60.8} &          54.8 &          52.8 &          52.1 \\
    +MASS  & \cellcolor{red!15}{51.9} & \cellcolor{red!15}{45.4} &          47.6 &                     61.0 &          54.9 &          54.2 &          52.5 \\
      +BT  &                     54.4 &                     47.4 &          50.0 &                     64.2 &          60.8 & \textbf{59.0} &          56.0 \\
\addlinespace
\textit{Mix+bal} & & & & & & & \\
+BART  & \cellcolor{red!15}{52.1} & \cellcolor{red!15}{44.6} &          47.9 & \cellcolor{red!15}{60.4} &          54.6 &          53.2 &          52.2 \\
+MASS  & \cellcolor{red!15}{52.5} & \cellcolor{red!15}{45.3} &          47.9 & \cellcolor{red!15}{60.6} &          55.6 &          54.9 &          52.9 \\
  +BT  &                     55.2 &                     48.8 & \textbf{50.5} &            \textbf{64.5} & \textbf{61.0} &          58.9 & \textbf{56.5} \\
\bottomrule
\end{tabular}

            \vspace{-4pt}
            \subcaption{chrF scores ($\uparrow$)}
            \label{tab:mmt-full-main-tico19-chrf}
        \end{minipage}\hfill
        \begin{minipage}{0.32\textwidth}
            \centering
            
\setlength{\tabcolsep}{1.5pt} %
\renewcommand{\arraystretch}{0.75} %
\scriptsize
\begin{tabular}{lrrrrrrr}
\toprule[1.5pt]
\multirow{2}[3]{*}{Model} &
\multicolumn{3}{c}{en$\rightarrow$xx} &
\multicolumn{3}{c}{xx$\rightarrow$en} &
\multicolumn{1}{c}{\multirow{2}[3]{*}{Mean}}
\\
\cmidrule(lr){2-4} \cmidrule(lr){5-7}
                        &                     High &                      Med &           Low &                     High &           Med & \multicolumn{2}{l}{Low} \\
\midrule
               parallel &                     80.3 &                     76.4 &          69.9 &                     83.4 &          79.2 &          73.1 &          76.7 \\
\addlinespace
\textit{Wiki} & & & & & & & \\
   +BART  & \cellcolor{red!15}{79.8} &                     76.6 &          70.9 &                     83.6 &          81.2 &          80.2 &          78.5 \\
   +MASS  & \cellcolor{red!15}{79.9} & \cellcolor{red!15}{75.6} &          70.9 &                     83.6 &          81.4 &          80.5 &          78.5 \\
     +BT  &                     81.1 &            \textbf{80.2} &          75.8 &                     84.7 &          83.0 &          80.5 &          80.7 \\
\addlinespace
\textit{Mix} & & & & & & & \\
    +BART  & \cellcolor{red!15}{78.9} & \cellcolor{red!15}{75.3} &          70.5 &                     83.4 &          81.2 &          80.0 &          78.1 \\
    +MASS  & \cellcolor{red!15}{79.2} & \cellcolor{red!15}{76.3} &          70.8 &                     83.6 &          81.3 &          81.0 &          78.5 \\
      +BT  &                     81.3 &                     79.3 &          76.9 &                     84.9 &          83.5 &          82.2 &          81.2 \\
\addlinespace
\textit{Mix+bal} & & & & & & & \\
+BART  & \cellcolor{red!15}{79.3} & \cellcolor{red!15}{75.4} &          70.6 & \cellcolor{red!15}{83.1} &          81.0 &          80.2 &          78.1 \\
+MASS  & \cellcolor{red!15}{79.5} & \cellcolor{red!15}{76.3} &          70.7 &                     83.4 &          81.8 &          81.5 &          78.7 \\
  +BT  &            \textbf{81.6} &                     80.1 & \textbf{77.3} &            \textbf{85.1} & \textbf{83.6} & \textbf{82.4} & \textbf{81.6} \\
\bottomrule
\end{tabular}

            \vspace{-4pt}
            \subcaption{COMET scores ($\uparrow$)}
            \label{tab:mmt-full-main-tico19-comet}
        \end{minipage}
        \caption{Results of the Transformer-Big models on the \textbf{TICO-19} (Medical) test set and grouped by the monolingual split that has been used for training BT and DAE. Cells in \colorbox{red!15}{red} indicate worse scores than the baseline.}
        \label{tab:mmt-full-main-tico19}
    \end{table*}

    \newpage

    \begin{table*}[t]
        \centering
        \begin{minipage}[t]{0.48\textwidth}
            \centering
            \includegraphics[width=\textwidth]{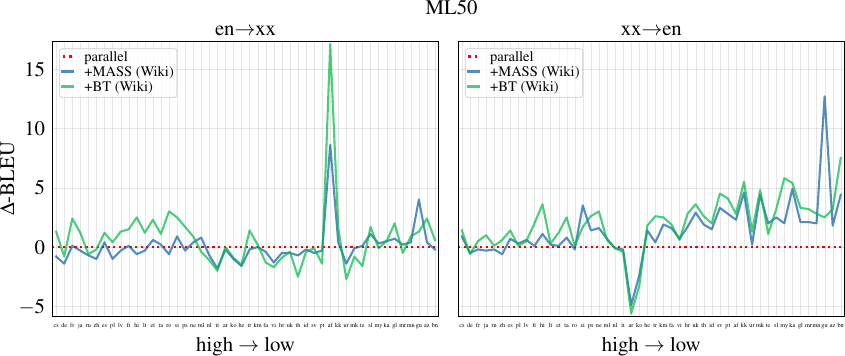}
            \includegraphics[width=\textwidth]{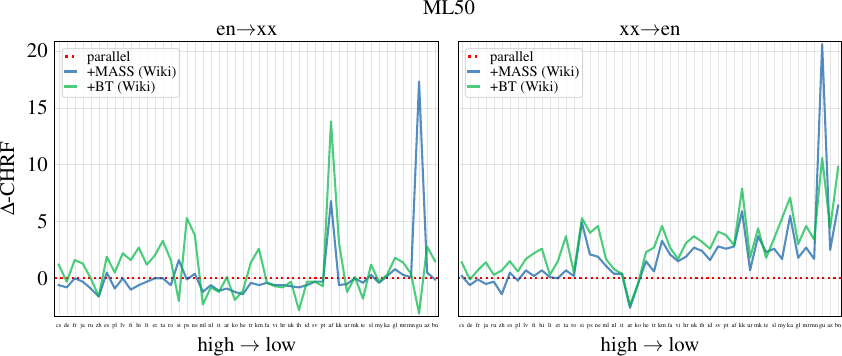}
            \includegraphics[width=\textwidth]{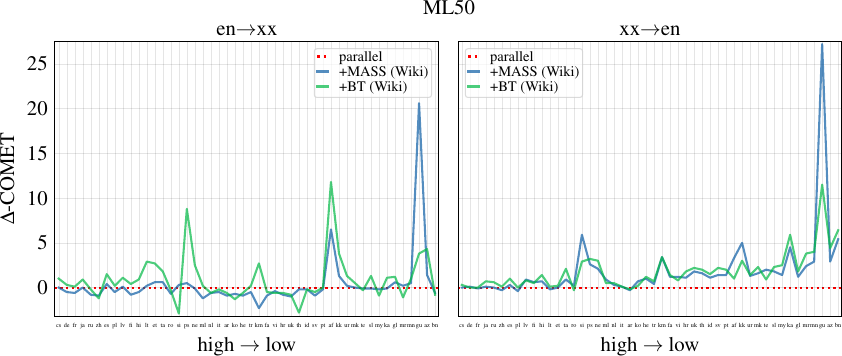}
            \subcaption{Results on ML50 test sets.}
            \label{tab:delta-scores-ml50}
            \par\vspace{16pt}
        \end{minipage}\hfill
        \begin{minipage}[t]{0.48\textwidth}
            \centering
            \includegraphics[width=\textwidth]{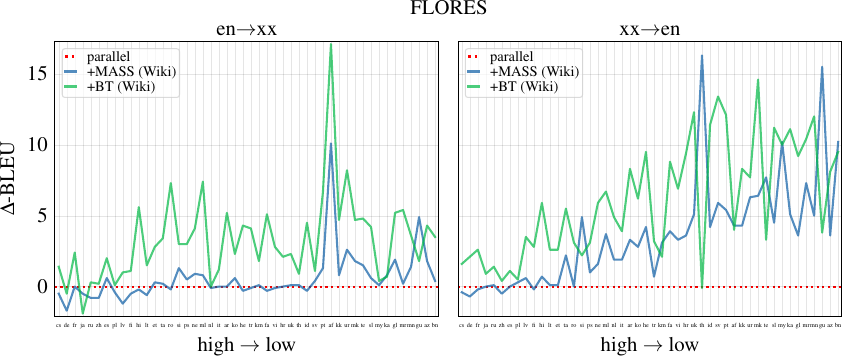}
            \includegraphics[width=\textwidth]{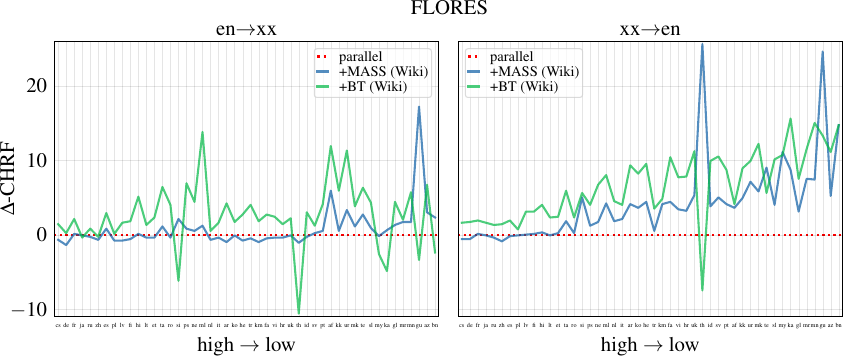}
            \includegraphics[width=\textwidth]{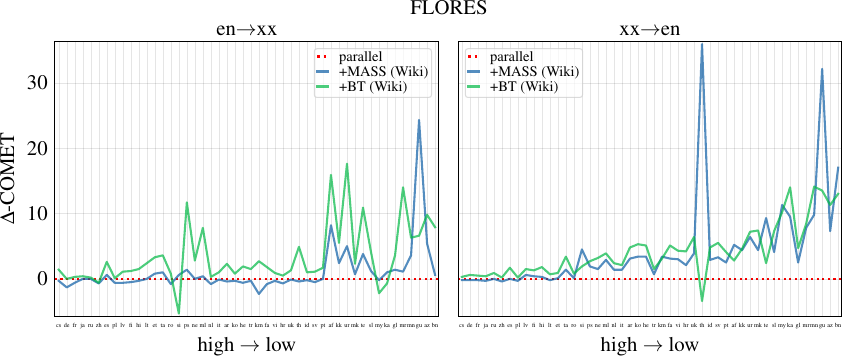}
            \subcaption{Results on FLORES (wiki) test sets.}
            \label{tab:delta-scores-flores}
            \par\vspace{16pt}
        \end{minipage}\hfill\par
        \begin{minipage}[t]{0.48\textwidth}
            \centering
            \includegraphics[width=\textwidth]{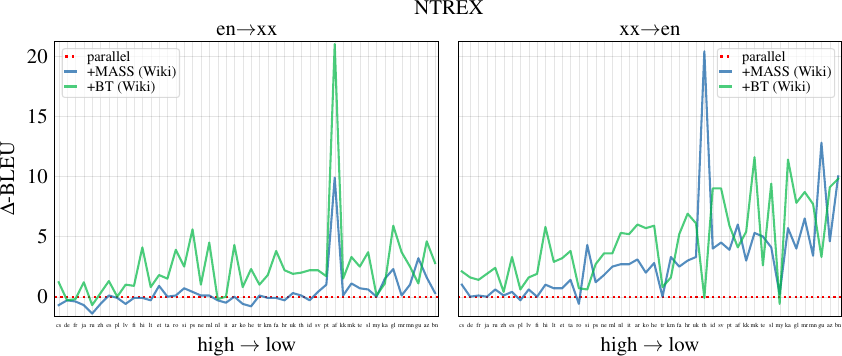}
            \includegraphics[width=\textwidth]{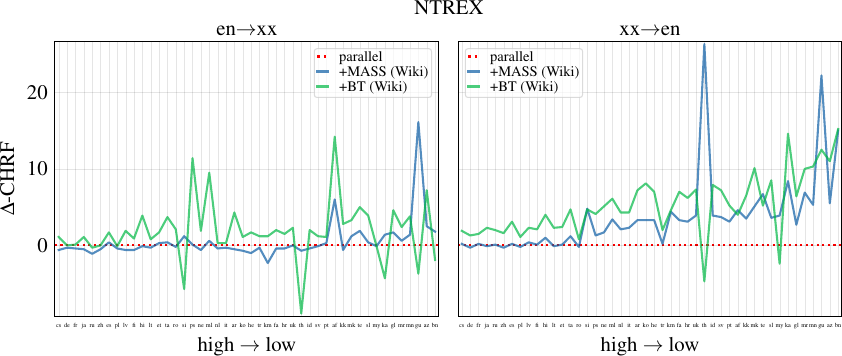}
            \includegraphics[width=\textwidth]{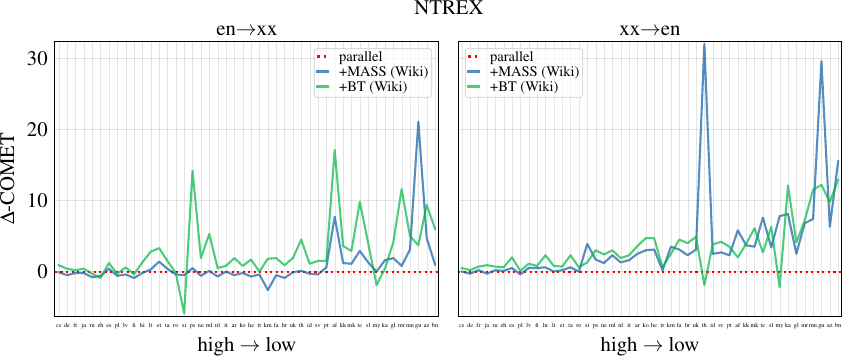}
            \subcaption{Results on NTREX (news) test sets.}
            \label{tab:delta-scores-ntrex}
            \par\vspace{16pt}
        \end{minipage}\hfill
        \begin{minipage}[t]{0.48\textwidth}
            \centering
            \includegraphics[width=\textwidth]{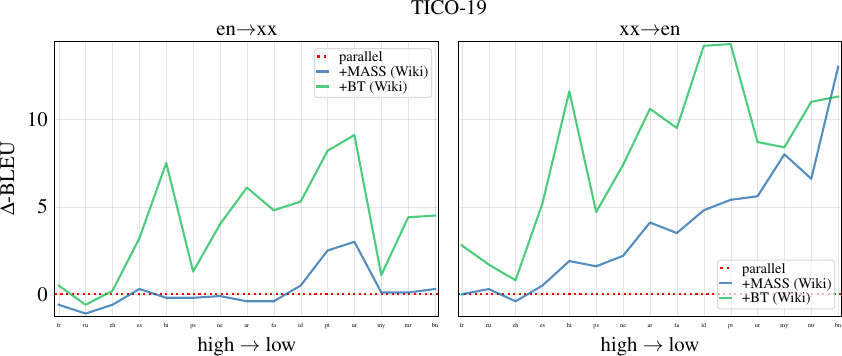}
            \includegraphics[width=\textwidth]{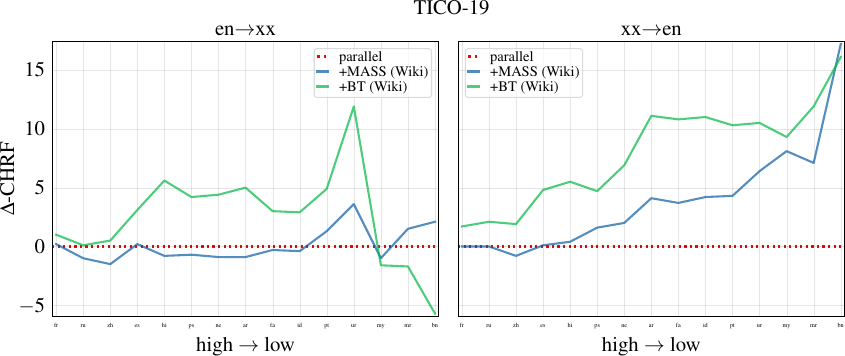}
            \includegraphics[width=\textwidth]{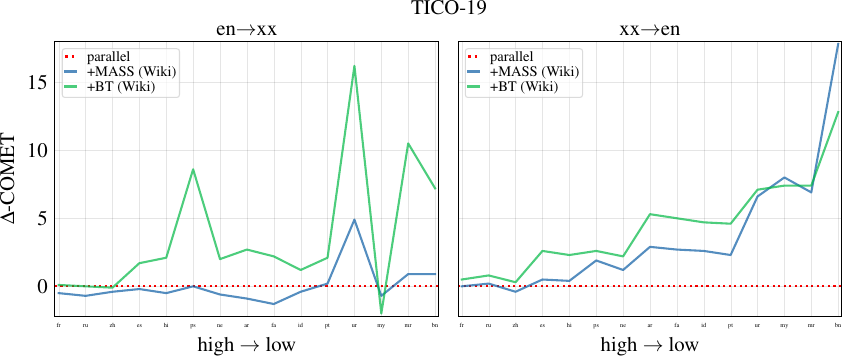}
            \subcaption{Results on TICO-19 (medical) test sets.}
            \label{tab:delta-scores-tico19}
            \par\vspace{16pt}
        \end{minipage}
        \caption{Score (BLEU, chrF, COMET) differences between each model and the parallel-only baseline (red dotted line) across test sets, for models with the Transformer-Big architecture (370M).}
        \label{fig:delta-scores-all}
    \end{table*}

    \newpage

    \newpage
    \clearpage

    \subsection{Denoising Autoencoding Objectives}
    \label{sec:app-dae}
    In this section, we extend the comparison of the two DAE objectives
    that is presented in Section~\ref{sec:ssl} by including the results across all metrics and monolingual splits.
    Specifically, Table~\ref{tab:mmt-full-ssl-mixbal} shows the results with the balanced mixed-domain monolingual split,
    Table~\ref{tab:mmt-full-ssl-mix} with the unbalanced mixed-domain monolingual split, and
    Table~\ref{tab:mmt-full-ssl-wiki} with the single-domain (Wikipedia) monolingual split.
    We observe that the differences are very small between models, but MASS outperforms BART by a small margin in most settings,
    similar to what is discussed in the main paper.

    \begin{table*}[t]
        \centering
        \begin{minipage}{0.32\textwidth}
            \centering
            
\setlength{\tabcolsep}{1.5pt} %
\renewcommand{\arraystretch}{1.0} %
\scriptsize
\begin{tabular}{lrrrrrrr}
\toprule[1.5pt]
\multirow{2}[3]{*}{Model} &
\multicolumn{3}{c}{en$\rightarrow$xx} &
\multicolumn{3}{c}{xx$\rightarrow$en} &
\multicolumn{1}{c}{\multirow{2}[3]{*}{Mean}}
\\
\cmidrule(lr){2-4} \cmidrule(lr){5-7}
               &          High &           Med &           Low &          High &           Med & \multicolumn{2}{l}{Low} \\
\midrule
\addlinespace\textit{FLORES} & & & & & & & \\
+BART  &          23.6 &          15.2 &          14.3 &          27.8 &          24.4 &          22.0 &          20.8 \\
+MASS  & \textbf{23.8} & \textbf{15.3} & \textbf{14.5} & \textbf{28.4} & \textbf{25.0} & \textbf{23.4} & \textbf{21.3} \\

\addlinespace\textit{NTREX} & & & & & & & \\
+BART  &          21.8 &          13.1 & \textbf{13.7} &          25.0 &          23.2 &          21.0 &          19.4 \\
+MASS  & \textbf{21.9} & \textbf{13.3} & \textbf{13.7} & \textbf{25.7} & \textbf{23.8} & \textbf{22.1} & \textbf{19.8} \\

\addlinespace\textit{ML50} & & & & & & & \\
+BART  & \textbf{22.1} & \textbf{16.8} &          21.3 &          26.8 &          26.1 &          28.1 &          23.5 \\
+MASS  & \textbf{22.1} & \textbf{16.8} & \textbf{21.5} & \textbf{27.2} & \textbf{26.6} & \textbf{28.8} & \textbf{23.8} \\

\addlinespace\textit{TICO-19} & & & & & & & \\
+BART  &          31.2 &          14.0 &          15.1 &          31.8 &          26.0 &          24.1 &          23.7 \\
+MASS  & \textbf{31.5} & \textbf{14.4} & \textbf{15.2} & \textbf{32.6} & \textbf{27.2} & \textbf{26.4} & \textbf{24.5} \\
\bottomrule
\end{tabular}

            \subcaption{BLEU scores ($\uparrow$)}
            \label{tab:mmt-full-ssl-mixbal-bleu}
        \end{minipage}\hfill
        \begin{minipage}{0.32\textwidth}
            \centering
            
\setlength{\tabcolsep}{1.5pt} %
\renewcommand{\arraystretch}{1.0} %
\scriptsize
\begin{tabular}{lrrrrrrr}
\toprule[1.5pt]
\multirow{2}[3]{*}{Model} &
\multicolumn{3}{c}{en$\rightarrow$xx} &
\multicolumn{3}{c}{xx$\rightarrow$en} &
\multicolumn{1}{c}{\multirow{2}[3]{*}{Mean}}
\\
\cmidrule(lr){2-4} \cmidrule(lr){5-7}
               &          High &           Med &           Low &          High &           Med & \multicolumn{2}{l}{Low} \\
\midrule
\addlinespace\textit{FLORES} & & & & & & & \\
+BART  &          49.4 &          45.3 &          46.5 &          56.4 &          52.9 &          50.4 &          49.9 \\
+MASS  & \textbf{49.7} & \textbf{45.8} & \textbf{46.8} & \textbf{56.8} & \textbf{53.3} & \textbf{51.6} & \textbf{50.4} \\

\addlinespace\textit{NTREX} & & & & & & & \\
+BART  &          47.4 &          42.9 &          44.1 &          54.4 &          51.8 &          48.2 &          47.9 \\
+MASS  & \textbf{47.6} & \textbf{43.3} & \textbf{44.3} & \textbf{54.6} & \textbf{52.1} & \textbf{49.3} & \textbf{48.3} \\

\addlinespace\textit{ML50} & & & & & & & \\
+BART  &          47.9 &          44.1 &          47.4 &          55.1 &          50.4 &          50.6 &          49.1 \\
+MASS  & \textbf{48.0} & \textbf{44.4} & \textbf{47.6} & \textbf{55.2} & \textbf{50.6} & \textbf{51.2} & \textbf{49.3} \\

\addlinespace\textit{TICO-19} & & & & & & & \\
+BART  &          52.1 &          44.6 & \textbf{47.9} &          60.4 &          54.6 &          53.2 &          52.2 \\
+MASS  & \textbf{52.5} & \textbf{45.3} & \textbf{47.9} & \textbf{60.6} & \textbf{55.6} & \textbf{54.9} & \textbf{52.9} \\
\bottomrule
\end{tabular}

            \subcaption{chrF scores ($\uparrow$)}
            \label{tab:mmt-full-ssl-mixbal-chrf}
        \end{minipage}\hfill
        \begin{minipage}{0.32\textwidth}
            \centering
            
\setlength{\tabcolsep}{1.5pt} %
\renewcommand{\arraystretch}{1.0} %
\scriptsize
\begin{tabular}{lrrrrrrr}
\toprule[1.5pt]
\multirow{2}[3]{*}{Model} &
\multicolumn{3}{c}{en$\rightarrow$xx} &
\multicolumn{3}{c}{xx$\rightarrow$en} &
\multicolumn{1}{c}{\multirow{2}[3]{*}{Mean}}
\\
\cmidrule(lr){2-4} \cmidrule(lr){5-7}
               &          High &           Med &           Low &          High &           Med & \multicolumn{2}{l}{Low} \\
\midrule
\addlinespace\textit{FLORES} & & & & & & & \\
+BART  &          81.9 &          79.5 &          73.4 &          83.3 &          81.0 &          78.5 &          79.3 \\
+MASS  & \textbf{82.3} & \textbf{80.0} & \textbf{74.0} & \textbf{83.7} & \textbf{81.6} & \textbf{79.7} & \textbf{79.9} \\

\addlinespace\textit{NTREX} & & & & & & & \\
+BART  &          78.2 &          76.3 &          71.9 &          81.8 &          80.5 &          75.9 &          77.2 \\
+MASS  & \textbf{78.6} & \textbf{76.8} & \textbf{72.3} & \textbf{82.2} & \textbf{81.0} & \textbf{77.2} & \textbf{77.8} \\

\addlinespace\textit{ML50} & & & & & & & \\
+BART  &          80.2 &          79.5 &          78.3 &          80.8 &          78.9 &          79.0 &          79.4 \\
+MASS  & \textbf{80.6} & \textbf{79.7} & \textbf{78.8} & \textbf{81.2} & \textbf{79.3} & \textbf{79.8} & \textbf{79.8} \\

\addlinespace\textit{TICO-19} & & & & & & & \\
+BART  &          79.3 &          75.4 &          70.6 &          83.1 &          81.0 &          80.2 &          78.1 \\
+MASS  & \textbf{79.5} & \textbf{76.3} & \textbf{70.7} & \textbf{83.4} & \textbf{81.8} & \textbf{81.5} & \textbf{78.7} \\
\bottomrule
\end{tabular}

            \subcaption{COMET scores ($\uparrow$)}
            \label{tab:mmt-full-ssl-mixbal-comet}
        \end{minipage}
        \caption{Comparison of the DAE objectives with models trained on the \textbf{balanced mixed-domain}.}
        \label{tab:mmt-full-ssl-mixbal}
    \end{table*}

    \begin{table*}[t]
        \centering
        \begin{minipage}{0.32\textwidth}
            \centering
            
\setlength{\tabcolsep}{1.5pt} %
\renewcommand{\arraystretch}{1.0} %
\scriptsize
\begin{tabular}{lrrrrrrr}
\toprule[1.5pt]
\multirow{2}[3]{*}{Model} &
\multicolumn{3}{c}{en$\rightarrow$xx} &
\multicolumn{3}{c}{xx$\rightarrow$en} &
\multicolumn{1}{c}{\multirow{2}[3]{*}{Mean}}
\\
\cmidrule(lr){2-4} \cmidrule(lr){5-7}
               &          High &           Med &           Low &          High &           Med & \multicolumn{2}{l}{Low} \\
\midrule
\addlinespace\textit{FLORES} & & & & & & & \\
+BART  &          23.4 &          15.0 & \textbf{14.2} &          27.9 &          24.7 &          22.6 &          20.9 \\
+MASS  & \textbf{23.5} & \textbf{15.2} &          14.1 & \textbf{28.5} & \textbf{24.8} & \textbf{23.0} & \textbf{21.1} \\

\addlinespace\textit{NTREX} & & & & & & & \\
+BART  &          21.6 &          13.0 &          13.6 &          25.4 &          23.4 &          21.5 &          19.5 \\
+MASS  & \textbf{21.7} & \textbf{13.1} & \textbf{13.7} & \textbf{26.0} & \textbf{23.9} & \textbf{22.1} & \textbf{19.8} \\

\addlinespace\textit{ML50} & & & & & & & \\
+BART  &          21.8 &          16.7 &          21.4 &          27.1 &          26.3 &          28.4 &          23.6 \\
+MASS  & \textbf{22.0} & \textbf{16.8} & \textbf{21.5} & \textbf{27.4} & \textbf{26.5} & \textbf{28.9} & \textbf{23.8} \\

\addlinespace\textit{TICO-19} & & & & & & & \\
+BART  &          30.5 &          13.9 &          14.9 &          32.5 &          26.6 &          24.2 &          23.7 \\
+MASS  & \textbf{31.1} & \textbf{14.3} & \textbf{15.2} & \textbf{33.0} & \textbf{26.9} & \textbf{25.6} & \textbf{24.3} \\
\bottomrule
\end{tabular}

            \subcaption{BLEU scores ($\uparrow$)}
            \label{tab:mmt-full-ssl-mix-bleu}
        \end{minipage}\hfill
        \begin{minipage}{0.32\textwidth}
            \centering
            
\setlength{\tabcolsep}{1.5pt} %
\renewcommand{\arraystretch}{1.0} %
\scriptsize
\begin{tabular}{lrrrrrrr}
\toprule[1.5pt]
\multirow{2}[3]{*}{Model} &
\multicolumn{3}{c}{en$\rightarrow$xx} &
\multicolumn{3}{c}{xx$\rightarrow$en} &
\multicolumn{1}{c}{\multirow{2}[3]{*}{Mean}}
\\
\cmidrule(lr){2-4} \cmidrule(lr){5-7}
               &          High &           Med &           Low &          High &           Med & \multicolumn{2}{l}{Low} \\
\midrule
\addlinespace\textit{FLORES} & & & & & & & \\
+BART  & \textbf{49.5} &          45.1 & \textbf{46.4} &          56.5 &          52.9 &          50.6 &          49.9 \\
+MASS  &          49.4 & \textbf{45.7} & \textbf{46.4} & \textbf{56.7} & \textbf{53.0} & \textbf{51.2} & \textbf{50.1} \\

\addlinespace\textit{NTREX} & & & & & & & \\
+BART  &          47.4 &          42.8 &          43.9 &          54.5 &          51.7 &          48.4 &          47.9 \\
+MASS  & \textbf{47.5} & \textbf{43.2} & \textbf{44.1} & \textbf{54.7} & \textbf{52.0} & \textbf{49.1} & \textbf{48.2} \\

\addlinespace\textit{ML50} & & & & & & & \\
+BART  &          47.7 &          44.0 &          47.4 & \textbf{55.3} & \textbf{50.4} &          50.8 &          49.1 \\
+MASS  & \textbf{47.9} & \textbf{44.4} & \textbf{47.5} & \textbf{55.3} & \textbf{50.4} & \textbf{51.2} & \textbf{49.3} \\

\addlinespace\textit{TICO-19} & & & & & & & \\
+BART  &          51.7 &          44.5 & \textbf{47.6} &          60.8 &          54.8 &          52.8 &          52.1 \\
+MASS  & \textbf{51.9} & \textbf{45.4} & \textbf{47.6} & \textbf{61.0} & \textbf{54.9} & \textbf{54.2} & \textbf{52.5} \\
\bottomrule
\end{tabular}

            \subcaption{chrF scores ($\uparrow$)}
            \label{tab:mmt-full-ssl-mix-chrf}
        \end{minipage}\hfill
        \begin{minipage}{0.32\textwidth}
            \centering
            
\setlength{\tabcolsep}{1.5pt} %
\renewcommand{\arraystretch}{1.0} %
\scriptsize
\begin{tabular}{lrrrrrrr}
\toprule[1.5pt]
\multirow{2}[3]{*}{Model} &
\multicolumn{3}{c}{en$\rightarrow$xx} &
\multicolumn{3}{c}{xx$\rightarrow$en} &
\multicolumn{1}{c}{\multirow{2}[3]{*}{Mean}}
\\
\cmidrule(lr){2-4} \cmidrule(lr){5-7}
               &          High &           Med &           Low &          High &           Med & \multicolumn{2}{l}{Low} \\
\midrule
\addlinespace\textit{FLORES} & & & & & & & \\
+BART  &          81.8 &          79.1 &          73.5 &          83.4 &          81.1 &          78.6 &          79.3 \\
+MASS  & \textbf{82.0} & \textbf{80.0} & \textbf{73.7} & \textbf{83.7} & \textbf{81.3} & \textbf{79.3} & \textbf{79.7} \\

\addlinespace\textit{NTREX} & & & & & & & \\
+BART  &          78.0 &          76.0 &          72.0 &          81.9 &          80.5 &          76.2 &          77.2 \\
+MASS  & \textbf{78.5} & \textbf{76.9} & \textbf{72.3} & \textbf{82.3} & \textbf{81.0} & \textbf{76.9} & \textbf{77.8} \\

\addlinespace\textit{ML50} & & & & & & & \\
+BART  &          80.0 &          79.2 &          78.7 &          80.9 &          79.0 &          79.3 &          79.4 \\
+MASS  & \textbf{80.5} & \textbf{79.9} & \textbf{78.8} & \textbf{81.2} & \textbf{79.3} & \textbf{79.8} & \textbf{79.8} \\

\addlinespace\textit{TICO-19} & & & & & & & \\
+BART  &          78.9 &          75.3 &          70.5 &          83.4 &          81.2 &          80.0 &          78.1 \\
+MASS  & \textbf{79.2} & \textbf{76.3} & \textbf{70.8} & \textbf{83.6} & \textbf{81.3} & \textbf{81.0} & \textbf{78.5} \\
\bottomrule
\end{tabular}

            \subcaption{COMET scores ($\uparrow$)}
            \label{tab:mmt-full-ssl-mix-comet}
        \end{minipage}
        \caption{Comparison of the DAE objectives with models trained on the \textbf{unbalanced mixed-domain}.}
        \label{tab:mmt-full-ssl-mix}
    \end{table*}

    \begin{table*}[t]
        \centering
        \begin{minipage}{0.32\textwidth}
            \centering
            
\setlength{\tabcolsep}{1.5pt} %
\renewcommand{\arraystretch}{1.0} %
\scriptsize
\begin{tabular}{lrrrrrrr}
\toprule[1.5pt]
\multirow{2}[3]{*}{Model} &
\multicolumn{3}{c}{en$\rightarrow$xx} &
\multicolumn{3}{c}{xx$\rightarrow$en} &
\multicolumn{1}{c}{\multirow{2}[3]{*}{Mean}}
\\
\cmidrule(lr){2-4} \cmidrule(lr){5-7}
               &          High &           Med &           Low &          High &           Med & \multicolumn{2}{l}{Low} \\
\midrule
\addlinespace\textit{FLORES} & & & & & & & \\
+BART  &          24.0 & \textbf{15.6} &          14.7 &          28.3 &          24.9 &          22.5 &          21.2 \\
+MASS  & \textbf{24.3} &          15.5 & \textbf{14.9} & \textbf{28.6} & \textbf{25.2} & \textbf{23.0} & \textbf{21.5} \\

\addlinespace\textit{NTREX} & & & & & & & \\
+BART  &          21.9 & \textbf{13.2} &          13.4 & \textbf{25.5} & \textbf{23.3} &          20.8 &          19.4 \\
+MASS  & \textbf{22.1} & \textbf{13.2} & \textbf{13.8} & \textbf{25.5} & \textbf{23.3} & \textbf{21.2} & \textbf{19.5} \\

\addlinespace\textit{ML50} & & & & & & & \\
+BART  &          22.0 & \textbf{16.9} & \textbf{21.3} &          27.0 & \textbf{26.6} &          27.9 &          23.6 \\
+MASS  & \textbf{22.1} & \textbf{16.9} & \textbf{21.3} & \textbf{27.1} &          26.5 & \textbf{28.5} & \textbf{23.7} \\

\addlinespace\textit{TICO-19} & & & & & & & \\
+BART  & \textbf{31.9} & \textbf{14.9} &          15.1 & \textbf{32.9} &          26.3 &          24.2 &          24.2 \\
+MASS  & \textbf{31.9} &          14.0 & \textbf{15.4} & \textbf{32.9} & \textbf{27.0} & \textbf{24.6} & \textbf{24.3} \\
\bottomrule
\end{tabular}

            \subcaption{BLEU scores ($\uparrow$)}
            \label{tab:mmt-full-ssl-wiki-bleu}
        \end{minipage}\hfill
        \begin{minipage}{0.32\textwidth}
            \centering
            
\setlength{\tabcolsep}{1.5pt} %
\renewcommand{\arraystretch}{1.0} %
\scriptsize
\begin{tabular}{lrrrrrrr}
\toprule[1.5pt]
\multirow{2}[3]{*}{Model} &
\multicolumn{3}{c}{en$\rightarrow$xx} &
\multicolumn{3}{c}{xx$\rightarrow$en} &
\multicolumn{1}{c}{\multirow{2}[3]{*}{Mean}}
\\
\cmidrule(lr){2-4} \cmidrule(lr){5-7}
               &          High &           Med &           Low &          High &           Med & \multicolumn{2}{l}{Low} \\
\midrule
\addlinespace\textit{FLORES} & & & & & & & \\
+BART  &          49.9 & \textbf{46.2} &          47.1 &          56.9 &          53.1 &          50.6 &          50.4 \\
+MASS  & \textbf{50.2} &          46.1 & \textbf{47.3} & \textbf{57.0} & \textbf{53.4} & \textbf{51.3} & \textbf{50.6} \\

\addlinespace\textit{NTREX} & & & & & & & \\
+BART  &          47.6 & \textbf{43.2} &          44.0 &          54.5 &          51.7 &          47.9 &          47.9 \\
+MASS  & \textbf{47.9} &          43.0 & \textbf{44.2} & \textbf{54.6} & \textbf{51.9} & \textbf{48.5} & \textbf{48.1} \\

\addlinespace\textit{ML50} & & & & & & & \\
+BART  &          47.9 & \textbf{44.6} &          47.4 &          55.1 &          50.5 &          50.4 &          49.1 \\
+MASS  & \textbf{48.1} &          44.5 & \textbf{47.5} & \textbf{55.2} & \textbf{50.6} & \textbf{50.9} & \textbf{49.3} \\

\addlinespace\textit{TICO-19} & & & & & & & \\
+BART  & \textbf{52.8} & \textbf{46.2} &          47.9 & \textbf{61.0} &          54.6 & \textbf{53.3} & \textbf{52.6} \\
+MASS  & \textbf{52.8} &          44.8 & \textbf{48.0} & \textbf{61.0} & \textbf{55.2} & \textbf{53.3} &          52.5 \\
\bottomrule
\end{tabular}

            \subcaption{chrF scores ($\uparrow$)}
            \label{tab:mmt-full-ssl-wiki-chrf}
        \end{minipage}\hfill
        \begin{minipage}{0.32\textwidth}
            \centering
            
\setlength{\tabcolsep}{1.5pt} %
\renewcommand{\arraystretch}{1.0} %
\scriptsize
\begin{tabular}{lrrrrrrr}
\toprule[1.5pt]
\multirow{2}[3]{*}{Model} &
\multicolumn{3}{c}{en$\rightarrow$xx} &
\multicolumn{3}{c}{xx$\rightarrow$en} &
\multicolumn{1}{c}{\multirow{2}[3]{*}{Mean}}
\\
\cmidrule(lr){2-4} \cmidrule(lr){5-7}
               &          High &           Med &           Low &          High &           Med & \multicolumn{2}{l}{Low} \\
\midrule
\addlinespace\textit{FLORES} & & & & & & & \\
+BART  &          82.5 & \textbf{80.5} &          74.4 & \textbf{83.9} &          81.5 &          78.9 &          80.0 \\
+MASS  & \textbf{82.8} &          80.3 & \textbf{74.6} & \textbf{83.9} & \textbf{81.7} & \textbf{79.6} & \textbf{80.2} \\

\addlinespace\textit{NTREX} & & & & & & & \\
+BART  &          78.3 & \textbf{76.7} &          72.0 &          82.1 &          80.6 &          75.5 &          77.3 \\
+MASS  & \textbf{78.8} &          76.4 & \textbf{72.3} & \textbf{82.3} & \textbf{80.8} & \textbf{76.3} & \textbf{77.6} \\

\addlinespace\textit{ML50} & & & & & & & \\
+BART  &          80.4 & \textbf{80.2} &          78.4 &          80.9 &          79.1 &          78.9 &          79.5 \\
+MASS  & \textbf{80.7} &          79.9 & \textbf{78.6} & \textbf{81.0} & \textbf{79.4} & \textbf{79.4} & \textbf{79.7} \\

\addlinespace\textit{TICO-19} & & & & & & & \\
+BART  &          79.8 & \textbf{76.6} & \textbf{70.9} & \textbf{83.6} &          81.2 &          80.2 & \textbf{78.5} \\
+MASS  & \textbf{79.9} &          75.6 & \textbf{70.9} & \textbf{83.6} & \textbf{81.4} & \textbf{80.5} & \textbf{78.5} \\
\bottomrule
\end{tabular}

            \subcaption{COMET scores ($\uparrow$)}
            \label{tab:mmt-full-ssl-wiki-comet}
        \end{minipage}
        \caption{Comparison of the DAE objectives with models trained on the \textbf{(Wikipedia) single-domain}.}
        \label{tab:mmt-full-ssl-wiki}
    \end{table*}

    \clearpage

    \subsection{Scaling}
    \label{sec:app-scaling}
    In this section, we report all of our results for the model scale analysis (\S\ref{sec:scale}).
    Tables~\ref{tab:mmt-full-scale-ml50},~\ref{tab:mmt-full-scale-flores},~\ref{tab:mmt-full-scale-ntrex},~\ref{tab:mmt-full-scale-tico19}
    show the results on the ML50, FLORES, NTREX and TICO19 test sets, respectively.
    For each test set, we report side-by-side the results from each evaluation metric.

\begin{figure*}[t]

        \begin{minipage}[t]{.32\linewidth}
            \centering

            \begin{subfigure}{\linewidth}
                \centering
                \includegraphics[width=\linewidth]{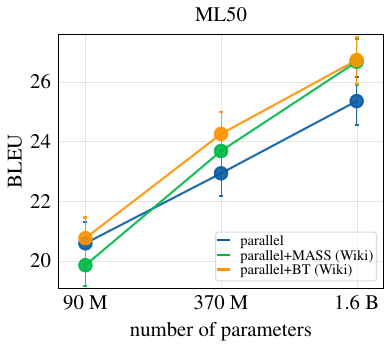}
            \end{subfigure}
            \hfill\par\vspace*{2pt}

            \begin{subfigure}{\linewidth}
                \centering
                \includegraphics[width=\linewidth]{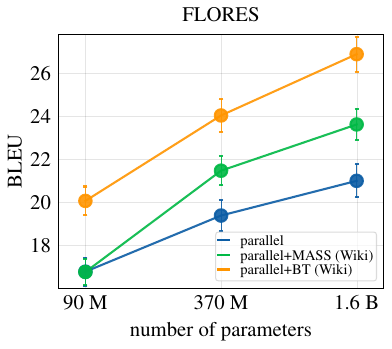}
            \end{subfigure}
            \hfill\par\vspace*{2pt}

            \begin{subfigure}{\linewidth}
                \centering
                \includegraphics[width=\linewidth]{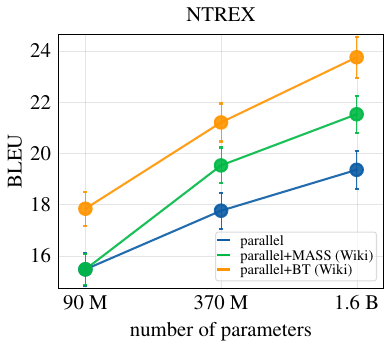}
            \end{subfigure}
            \hfill\par\vspace*{2pt}

            \begin{subfigure}{\linewidth}
                \centering
                \includegraphics[width=\linewidth]{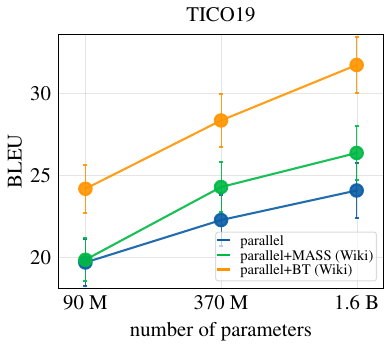}
            \end{subfigure}
            \hfill\par\vspace*{2pt}

            \caption{Average BLEU scores across model scales.
            }
            \label{fig:scale-line-bleu-all}
        \end{minipage}
        \hfill
        \begin{minipage}[t]{.32\linewidth}
            \centering

            \begin{subfigure}{\linewidth}
                \centering
                \includegraphics[width=\linewidth]{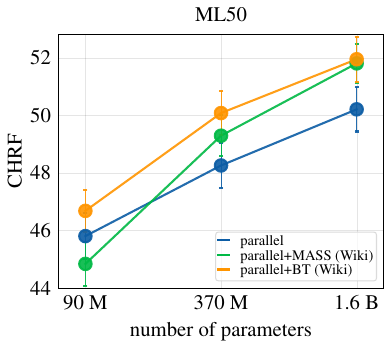}
            \end{subfigure}
            \hfill\par\vspace*{2pt}

            \begin{subfigure}{\linewidth}
                \centering
                \includegraphics[width=\linewidth]{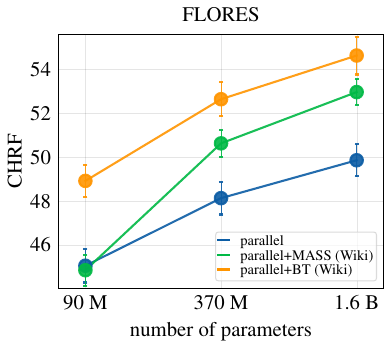}
            \end{subfigure}
            \hfill\par\vspace*{2pt}

            \begin{subfigure}{\linewidth}
                \centering
                \includegraphics[width=\linewidth]{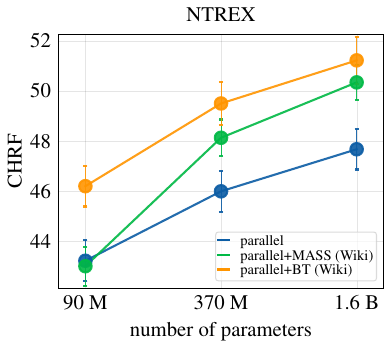}
            \end{subfigure}
            \hfill\par\vspace*{2pt}

            \begin{subfigure}{\linewidth}
                \centering
                \includegraphics[width=\linewidth]{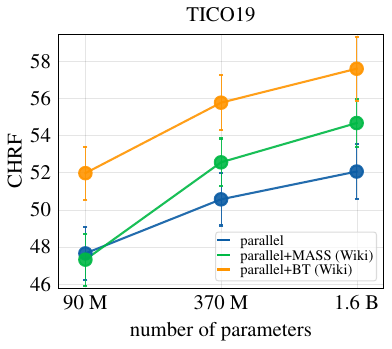}
            \end{subfigure}
            \hfill\par\vspace*{2pt}

            \caption{Average ChrF scores across model scales.
            }
            \label{fig:scale-line-chrf-all}
        \end{minipage}
        \hfill
        \begin{minipage}[t]{.32\linewidth}
            \centering

            \begin{subfigure}{\linewidth}
                \centering
                \includegraphics[width=\linewidth]{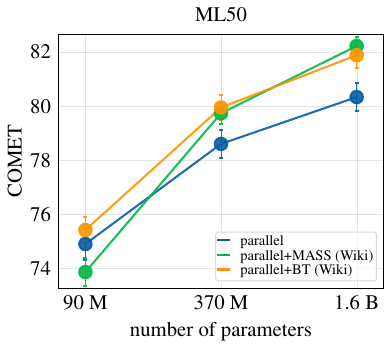}
            \end{subfigure}
            \hfill\par\vspace*{2pt}

            \begin{subfigure}{\linewidth}
                \centering
                \includegraphics[width=\linewidth]{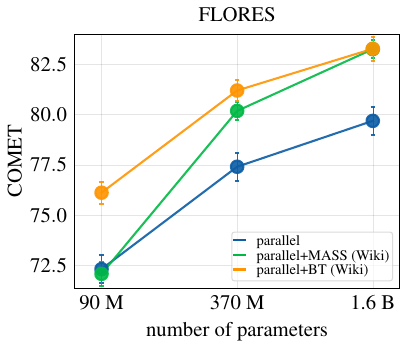}
            \end{subfigure}
            \hfill\par\vspace*{2pt}

            \begin{subfigure}{\linewidth}
                \centering
                \includegraphics[width=\linewidth]{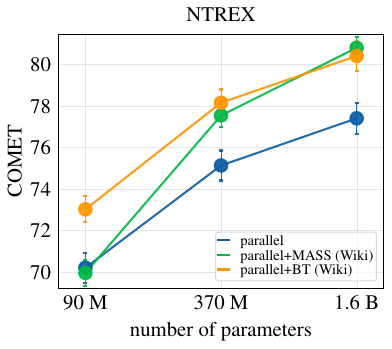}
            \end{subfigure}
            \hfill\par\vspace*{2pt}

            \begin{subfigure}{\linewidth}
                \centering
                \includegraphics[width=\linewidth]{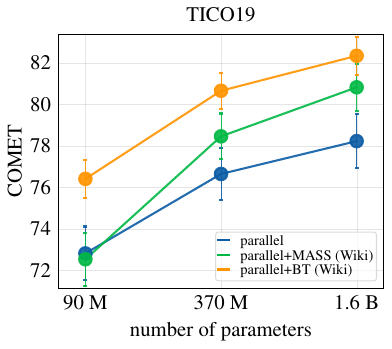}
            \end{subfigure}
            \hfill\par\vspace*{2pt}

            \caption{Average COMET scores across model scales.
            }
            \label{fig:scale-line-comet-all}
        \end{minipage}

    \end{figure*}

    \paragraph{Model Averages per Scale}
    As it is not easy to extract meaningful patterns from the results in table format,
    we also plot the corresponding line plots with the average score of each method per model scale
    across metrics,
    in Figure~\ref{fig:scale-line-bleu-all} (BLEU),
    Figure~\ref{fig:scale-line-chrf-all} (chrF),
    and Figure~\ref{fig:scale-line-comet-all} (COMET).
    We observe that the trends are overall the same across both metrics.
    All metrics agree that at small scales, MASS fails to outperform the baseline but becomes much more effective,
    compared to the baseline, as the scale increases.
    This further supports the findings discussed in the main paper.

    However, we discover that metrics disagree with each other about the \textit{degree} that scale benefits DAE/MASS.
    Specifically, we see that according to BLEU, DAE at the 1.6B scale is competitive with BT only on the ML50 test set,
    whereas chrF (middle column) and COMET (right column) suggest that DAE becomes much stronger with scale.
    In particular, according to COMET, at the 1.6B scale, MASS matches or outperforms BT on most test sets.

    \paragraph{Model Averages per Resource-Level}
    For completeness, we also include the plots with the scaling patterns of each model
    across resource levels and translation directions,
    in Figure~\ref{fig:app-scale-analysis-methods-bleu-detailed} (BLEU; left column),
    Figure~\ref{fig:app-scale-analysis-methods-chrf-detailed} (chrF; middle column),
    Figure~\ref{fig:app-scale-analysis-methods-comet-detailed} (COMET; right column).
    Overall, the results are consistent across metrics and test sets and the discussion in the main paper still holds.

    However, we do discover one interesting discrepancy, which potentially relates to the observations of the previous paragraph.
    Specifically, in the chrF plots we see that BT in en$\rightarrow$xx low-resource settings (bottom-left plot per test set)
    tends to become less effective than the parallel baseline in all test sets except for ML50.
    Recall that ML50 is the most distant test set with respect to the (Wikipedia) monolingual data.
    We do not have a reliable explanation for this observation.

    \begin{table*}[bt]
        \centering
        \begin{minipage}{0.32\textwidth}
            \centering
            
\setlength{\tabcolsep}{1.5pt} %
\renewcommand{\arraystretch}{0.8} %
\scriptsize
\begin{tabular}{lrrrrrrr}
\toprule[1.5pt]
\multirow{2}[3]{*}{Model} &
\multicolumn{3}{c}{en$\rightarrow$xx} &
\multicolumn{3}{c}{xx$\rightarrow$en} &
\multicolumn{1}{c}{\multirow{2}[3]{*}{Mean}}
\\
\cmidrule(lr){2-4} \cmidrule(lr){5-7}
         &          High &           Med &           Low &          High &           Med & \multicolumn{2}{l}{Low} \\
\midrule
\addlinespace\textit{Base} & & & & & & & \\
parallel &          18.8 &          14.8 & \textbf{18.3} &          23.7 & \textbf{23.2} &          24.9 &          20.6 \\
   +MASS &          18.0 &          13.9 &          17.4 &          23.0 &          22.3 &          24.7 &          19.9 \\
     +BT & \textbf{19.7} & \textbf{14.9} &          17.7 & \textbf{24.3} &          23.0 & \textbf{25.5} & \textbf{20.8} \\

\addlinespace\textit{Big} & & & & & & & \\
parallel &          22.5 & \textbf{17.3} &          20.6 &          26.9 &          25.9 &          25.3 &          22.9 \\
   +MASS &          22.1 &          16.9 &          21.3 &          27.1 &          26.5 &          28.5 &          23.7 \\
     +BT & \textbf{23.6} & \textbf{17.3} & \textbf{21.6} & \textbf{27.8} & \textbf{26.9} & \textbf{28.9} & \textbf{24.3} \\

\addlinespace\textit{XL} & & & & & & & \\
parallel &          25.2 &          18.4 &          21.4 &          30.1 &          29.7 &          28.4 &          25.4 \\
   +MASS &          25.7 &          19.0 &          23.1 & \textbf{31.1} & \textbf{30.8} & \textbf{31.1} & \textbf{26.7} \\
     +BT & \textbf{26.5} & \textbf{19.3} & \textbf{23.2} &          31.0 &          30.6 &          30.7 & \textbf{26.7} \\
\bottomrule
\end{tabular}

            \subcaption{BLEU scores ($\uparrow$)}
            \label{tab:mmt-full-scale-ml50-bleu}
        \end{minipage}\hfill
        \begin{minipage}{0.32\textwidth}
            \centering
            
\setlength{\tabcolsep}{1.5pt} %
\renewcommand{\arraystretch}{0.8} %
\scriptsize
\begin{tabular}{lrrrrrrr}
\toprule[1.5pt]
\multirow{2}[3]{*}{Model} &
\multicolumn{3}{c}{en$\rightarrow$xx} &
\multicolumn{3}{c}{xx$\rightarrow$en} &
\multicolumn{1}{c}{\multirow{2}[3]{*}{Mean}}
\\
\cmidrule(lr){2-4} \cmidrule(lr){5-7}
         &          High &           Med &           Low &          High &           Med & \multicolumn{2}{l}{Low} \\
\midrule
\addlinespace\textit{Base} & & & & & & & \\
parallel &          44.9 &          41.3 &          43.3 &          52.3 &          47.4 &          47.0 &          45.8 \\
   +MASS &          43.9 &          39.9 &          42.1 &          51.3 &          46.3 &          46.9 &          44.9 \\
     +BT & \textbf{46.0} & \textbf{42.3} & \textbf{43.4} & \textbf{53.2} & \textbf{48.1} & \textbf{48.6} & \textbf{46.7} \\

\addlinespace\textit{Big} & & & & & & & \\
parallel &          48.6 &          45.0 &          46.3 &          55.3 &          49.3 &          47.0 &          48.3 \\
   +MASS &          48.1 &          44.5 & \textbf{47.5} &          55.2 &          50.6 &          50.9 &          49.3 \\
     +BT & \textbf{49.7} & \textbf{45.4} &          47.2 & \textbf{56.4} & \textbf{51.6} & \textbf{51.7} & \textbf{50.1} \\

\addlinespace\textit{XL} & & & & & & & \\
parallel &          50.9 &          46.2 &          46.7 &          57.8 &          52.1 &          49.7 &          50.2 \\
   +MASS &          51.2 &          46.9 & \textbf{49.3} &          58.6 &          53.4 &          52.9 &          51.8 \\
     +BT & \textbf{52.3} & \textbf{47.3} &          48.2 & \textbf{58.8} & \textbf{54.0} & \textbf{53.0} & \textbf{52.0} \\
\bottomrule
\end{tabular}

            \subcaption{chrF scores ($\uparrow$)}
            \label{tab:mmt-full-scale-ml50-chrf}
        \end{minipage}\hfill
        \begin{minipage}{0.32\textwidth}
            \centering
            
\setlength{\tabcolsep}{1.5pt} %
\renewcommand{\arraystretch}{0.8} %
\scriptsize
\begin{tabular}{lrrrrrrr}
\toprule[1.5pt]
\multirow{2}[3]{*}{Model} &
\multicolumn{3}{c}{en$\rightarrow$xx} &
\multicolumn{3}{c}{xx$\rightarrow$en} &
\multicolumn{1}{c}{\multirow{2}[3]{*}{Mean}}
\\
\cmidrule(lr){2-4} \cmidrule(lr){5-7}
         &          High &           Med &           Low &          High &           Med & \multicolumn{2}{l}{Low} \\
\midrule
\addlinespace\textit{Base} & & & & & & & \\
parallel &          75.0 &          75.4 & \textbf{73.1} &          76.9 &          75.3 &          74.3 &          74.9 \\
   +MASS &          73.2 &          73.2 &          71.9 &          76.0 &          74.4 &          74.8 &          73.9 \\
     +BT & \textbf{75.6} & \textbf{76.2} &          72.9 & \textbf{77.3} & \textbf{75.4} & \textbf{75.7} & \textbf{75.4} \\

\addlinespace\textit{Big} & & & & & & & \\
parallel &          80.9 &          80.5 &          77.0 &          80.8 &          78.0 &          75.6 &          78.6 \\
   +MASS &          80.7 &          79.9 & \textbf{78.6} &          81.0 & \textbf{79.4} & \textbf{79.4} &          79.7 \\
     +BT & \textbf{81.8} & \textbf{80.9} &          78.3 & \textbf{81.3} & \textbf{79.4} &          78.9 & \textbf{80.0} \\

\addlinespace\textit{XL} & & & & & & & \\
parallel &          83.7 &          82.2 &          77.7 &          83.4 &          79.8 &          77.1 &          80.3 \\
   +MASS &          84.0 &          82.9 & \textbf{81.1} & \textbf{84.1} &          81.1 & \textbf{81.0} & \textbf{82.2} \\
     +BT & \textbf{84.6} & \textbf{83.0} &          79.6 &          83.8 & \textbf{81.5} &          80.1 &          81.9 \\
\bottomrule
\end{tabular}

            \subcaption{COMET scores ($\uparrow$)}
            \label{tab:mmt-full-scale-ml50-comet}
        \end{minipage}
        \caption{Results of all methods across different model scales evaluated on the \textbf{ML50} (mixed-domain) test set. The BT and DAE models have used the (Wikipedia) single-domain monolingual split.}
        \label{tab:mmt-full-scale-ml50}
    \end{table*}

    \begin{table*}[bt]
        \centering
        \begin{minipage}{0.32\textwidth}
            \centering
            
\setlength{\tabcolsep}{1.5pt} %
\renewcommand{\arraystretch}{0.8} %
\scriptsize
\begin{tabular}{lrrrrrrr}
\toprule[1.5pt]
\multirow{2}[3]{*}{Model} &
\multicolumn{3}{c}{en$\rightarrow$xx} &
\multicolumn{3}{c}{xx$\rightarrow$en} &
\multicolumn{1}{c}{\multirow{2}[3]{*}{Mean}}
\\
\cmidrule(lr){2-4} \cmidrule(lr){5-7}
         &          High &           Med &           Low &          High &           Med & \multicolumn{2}{l}{Low} \\
\midrule
\addlinespace\textit{Base} & & & & & & & \\
parallel &          20.0 &          12.3 &          10.8 &          24.1 &          20.2 &          16.0 &          16.8 \\
   +MASS &          19.0 &          11.7 &          11.3 &          23.5 &          19.6 &          17.9 &          16.8 \\
     +BT & \textbf{21.1} & \textbf{15.3} & \textbf{14.9} & \textbf{26.4} & \textbf{23.6} & \textbf{21.1} & \textbf{20.1} \\

\addlinespace\textit{Big} & & & & & & & \\
parallel &          24.8 &          15.3 &          13.2 &          28.6 &          22.4 &          16.0 &          19.4 \\
   +MASS &          24.3 &          15.5 &          14.9 &          28.6 &          25.2 &          23.0 &          21.5 \\
     +BT & \textbf{26.0} & \textbf{18.8} & \textbf{17.7} & \textbf{30.8} & \textbf{28.4} & \textbf{24.9} & \textbf{24.1} \\

\addlinespace\textit{XL} & & & & & & & \\
parallel &          27.6 &          17.1 &          13.8 &          32.3 &          23.1 &          17.0 &          21.0 \\
   +MASS &          28.3 &          18.4 &          16.2 &          33.5 &          25.6 &          23.5 &          23.7 \\
     +BT & \textbf{29.6} & \textbf{21.1} & \textbf{19.2} & \textbf{34.4} & \textbf{32.3} & \textbf{27.5} & \textbf{26.9} \\
\bottomrule
\end{tabular}

            \subcaption{BLEU scores ($\uparrow$)}
            \label{tab:mmt-full-scale-flores-bleu}
        \end{minipage}\hfill
        \begin{minipage}{0.32\textwidth}
            \centering
            
\setlength{\tabcolsep}{1.5pt} %
\renewcommand{\arraystretch}{0.8} %
\scriptsize
\begin{tabular}{lrrrrrrr}
\toprule[1.5pt]
\multirow{2}[3]{*}{Model} &
\multicolumn{3}{c}{en$\rightarrow$xx} &
\multicolumn{3}{c}{xx$\rightarrow$en} &
\multicolumn{1}{c}{\multirow{2}[3]{*}{Mean}}
\\
\cmidrule(lr){2-4} \cmidrule(lr){5-7}
         &          High &           Med &           Low &          High &           Med & \multicolumn{2}{l}{Low} \\
\midrule
\addlinespace\textit{Base} & & & & & & & \\
parallel &          46.2 &          41.1 &          40.9 &          53.5 &          48.8 &          42.6 &          45.1 \\
   +MASS &          45.1 &          39.4 &          40.2 &          52.6 &          48.0 &          45.9 &          44.9 \\
     +BT & \textbf{47.9} & \textbf{45.2} & \textbf{44.5} & \textbf{55.8} & \textbf{53.0} & \textbf{48.8} & \textbf{48.9} \\

\addlinespace\textit{Big} & & & & & & & \\
parallel &          50.6 &          46.1 &          45.0 &          57.1 &          50.4 &          42.6 &          48.2 \\
   +MASS &          50.2 &          46.1 &          47.3 &          57.0 &          53.4 &          51.3 &          50.6 \\
     +BT & \textbf{52.1} & \textbf{49.3} & \textbf{47.6} & \textbf{59.2} & \textbf{57.1} & \textbf{52.2} & \textbf{52.7} \\

\addlinespace\textit{XL} & & & & & & & \\
parallel &          53.0 &          47.9 &          45.6 &          60.0 &          51.4 &          44.9 &          49.9 \\
   +MASS &          53.5 &          49.5 & \textbf{49.5} &          60.8 &          54.1 &          52.6 &          53.0 \\
     +BT & \textbf{54.8} & \textbf{51.5} &          47.2 & \textbf{62.0} & \textbf{60.2} & \textbf{54.4} & \textbf{54.6} \\
\bottomrule
\end{tabular}

            \subcaption{chrF scores ($\uparrow$)}
            \label{tab:mmt-full-scale-flores-chrf}
        \end{minipage}\hfill
        \begin{minipage}{0.32\textwidth}
            \centering
            
\setlength{\tabcolsep}{1.5pt} %
\renewcommand{\arraystretch}{0.8} %
\scriptsize
\begin{tabular}{lrrrrrrr}
\toprule[1.5pt]
\multirow{2}[3]{*}{Model} &
\multicolumn{3}{c}{en$\rightarrow$xx} &
\multicolumn{3}{c}{xx$\rightarrow$en} &
\multicolumn{1}{c}{\multirow{2}[3]{*}{Mean}}
\\
\cmidrule(lr){2-4} \cmidrule(lr){5-7}
         &          High &           Med &           Low &          High &           Med & \multicolumn{2}{l}{Low} \\
\midrule
\addlinespace\textit{Base} & & & & & & & \\
parallel &          76.9 &          73.1 &          64.9 &          79.7 &          75.7 &          67.1 &          72.4 \\
   +MASS &          74.9 &          70.0 &          64.2 &          78.9 &          75.2 &          72.1 &          72.1 \\
     +BT & \textbf{78.4} & \textbf{76.9} & \textbf{71.4} & \textbf{81.1} & \textbf{78.3} & \textbf{72.8} & \textbf{76.1} \\

\addlinespace\textit{Big} & & & & & & & \\
parallel &          83.0 &          80.5 &          71.4 &          84.0 &          79.3 &          69.9 &          77.4 \\
   +MASS &          82.8 &          80.3 &          74.6 &          83.9 &          81.7 & \textbf{79.6} &          80.2 \\
     +BT & \textbf{84.1} & \textbf{82.7} & \textbf{77.5} & \textbf{84.8} & \textbf{82.9} &          77.2 & \textbf{81.2} \\

\addlinespace\textit{XL} & & & & & & & \\
parallel &          85.6 &          83.1 &          72.7 &          86.3 &          81.0 &          73.2 &          79.7 \\
   +MASS &          86.1 &          84.4 &          78.1 & \textbf{86.9} &          83.4 & \textbf{82.5} & \textbf{83.3} \\
     +BT & \textbf{86.7} & \textbf{84.8} & \textbf{78.4} & \textbf{86.9} & \textbf{85.6} &          79.4 & \textbf{83.3} \\
\bottomrule
\end{tabular}

            \subcaption{COMET scores ($\uparrow$)}
            \label{tab:mmt-full-scale-flores-comet}
        \end{minipage}
        \caption{Results of all methods across different model scales evaluated on the \textbf{FLORES} (Wikipedia) test set. The BT and DAE models have used the (Wikipedia) single-domain monolingual split.}
        \label{tab:mmt-full-scale-flores}
    \end{table*}

    \begin{table*}[bt]
        \centering
        \begin{minipage}{0.32\textwidth}
            \centering
            
\setlength{\tabcolsep}{1.5pt} %
\renewcommand{\arraystretch}{0.8} %
\scriptsize
\begin{tabular}{lrrrrrrr}
\toprule[1.5pt]
\multirow{2}[3]{*}{Model} &
\multicolumn{3}{c}{en$\rightarrow$xx} &
\multicolumn{3}{c}{xx$\rightarrow$en} &
\multicolumn{1}{c}{\multirow{2}[3]{*}{Mean}}
\\
\cmidrule(lr){2-4} \cmidrule(lr){5-7}
         &          High &           Med &           Low &          High &           Med & \multicolumn{2}{l}{Low} \\
\midrule
\addlinespace\textit{Base} & & & & & & & \\
parallel &          18.5 &          10.7 &          10.5 &          21.5 &          18.7 &          15.0 &          15.5 \\
   +MASS &          17.8 &          10.2 &          10.8 &          20.9 &          18.2 &          16.7 &          15.5 \\
     +BT & \textbf{19.5} & \textbf{12.7} & \textbf{13.6} & \textbf{23.2} & \textbf{20.9} & \textbf{18.6} & \textbf{17.9} \\

\addlinespace\textit{Big} & & & & & & & \\
parallel &          22.4 &          13.2 &          12.4 &          25.1 &          21.1 &          15.1 &          17.8 \\
   +MASS &          22.1 &          13.2 &          13.8 &          25.5 &          23.3 &          21.2 &          19.5 \\
     +BT & \textbf{23.3} & \textbf{15.5} & \textbf{16.0} & \textbf{27.4} & \textbf{25.1} & \textbf{21.8} & \textbf{21.2} \\

\addlinespace\textit{XL} & & & & & & & \\
parallel &          24.6 &          14.4 &          13.0 &          29.2 &          22.2 &          16.1 &          19.4 \\
   +MASS &          25.1 &          15.6 &          15.0 &          29.9 &          24.5 &          21.7 &          21.6 \\
     +BT & \textbf{26.0} & \textbf{17.5} & \textbf{17.0} & \textbf{31.1} & \textbf{28.8} & \textbf{24.3} & \textbf{23.8} \\
\bottomrule
\end{tabular}

            \subcaption{BLEU scores ($\uparrow$)}
            \label{tab:mmt-full-scale-ntrex-bleu}
        \end{minipage}\hfill
        \begin{minipage}{0.32\textwidth}
            \centering
            
\setlength{\tabcolsep}{1.5pt} %
\renewcommand{\arraystretch}{0.8} %
\scriptsize
\begin{tabular}{lrrrrrrr}
\toprule[1.5pt]
\multirow{2}[3]{*}{Model} &
\multicolumn{3}{c}{en$\rightarrow$xx} &
\multicolumn{3}{c}{xx$\rightarrow$en} &
\multicolumn{1}{c}{\multirow{2}[3]{*}{Mean}}
\\
\cmidrule(lr){2-4} \cmidrule(lr){5-7}
         &          High &           Med &           Low &          High &           Med & \multicolumn{2}{l}{Low} \\
\midrule
\addlinespace\textit{Base} & & & & & & & \\
parallel &          44.4 &          38.9 &          38.5 &          51.6 &          47.5 &          40.6 &          43.2 \\
   +MASS &          43.5 &          37.4 &          37.8 &          50.7 &          46.8 &          43.6 &          43.0 \\
     +BT & \textbf{45.7} & \textbf{41.9} & \textbf{41.5} & \textbf{53.5} & \textbf{50.9} & \textbf{45.3} & \textbf{46.2} \\

\addlinespace\textit{Big} & & & & & & & \\
parallel &          48.3 &          43.3 &          42.2 &          54.5 &          49.3 &          40.8 &          46.0 \\
   +MASS &          47.9 &          43.0 & \textbf{44.2} &          54.6 &          51.9 & \textbf{48.5} &          48.1 \\
     +BT & \textbf{49.2} & \textbf{45.7} & \textbf{44.2} & \textbf{56.7} & \textbf{54.6} &          48.3 & \textbf{49.5} \\

\addlinespace\textit{XL} & & & & & & & \\
parallel &          50.3 &          44.7 &          42.8 &          57.5 &          50.7 &          43.0 &          47.7 \\
   +MASS &          50.6 &          46.1 & \textbf{46.4} &          58.0 &          52.9 &          49.8 &          50.4 \\
     +BT & \textbf{51.7} & \textbf{47.7} &          43.1 & \textbf{59.4} & \textbf{57.5} & \textbf{50.2} & \textbf{51.2} \\
\bottomrule
\end{tabular}

            \subcaption{chrF scores ($\uparrow$)}
            \label{tab:mmt-full-scale-ntrex-chrf}
        \end{minipage}\hfill
        \begin{minipage}{0.32\textwidth}
            \centering
            
\setlength{\tabcolsep}{1.5pt} %
\renewcommand{\arraystretch}{0.8} %
\scriptsize
\begin{tabular}{lrrrrrrr}
\toprule[1.5pt]
\multirow{2}[3]{*}{Model} &
\multicolumn{3}{c}{en$\rightarrow$xx} &
\multicolumn{3}{c}{xx$\rightarrow$en} &
\multicolumn{1}{c}{\multirow{2}[3]{*}{Mean}}
\\
\cmidrule(lr){2-4} \cmidrule(lr){5-7}
         &          High &           Med &           Low &          High &           Med & \multicolumn{2}{l}{Low} \\
\midrule
\addlinespace\textit{Base} & & & & & & & \\
parallel &          72.6 &          69.8 &          63.2 &          78.2 &          75.1 &          65.0 &          70.2 \\
   +MASS &          70.8 &          67.0 &          62.6 &          77.3 &          74.6 & \textbf{69.5} &          70.0 \\
     +BT & \textbf{73.2} & \textbf{72.4} & \textbf{68.3} & \textbf{79.2} & \textbf{77.2} & \textbf{69.5} & \textbf{73.0} \\

\addlinespace\textit{Big} & & & & & & & \\
parallel &          79.0 &          76.9 &          69.4 &          82.1 &          78.7 &          67.6 &          75.2 \\
   +MASS &          78.8 &          76.4 &          72.3 &          82.3 &          80.8 & \textbf{76.3} &          77.6 \\
     +BT & \textbf{79.7} & \textbf{78.7} & \textbf{74.4} & \textbf{83.0} & \textbf{81.6} &          73.4 & \textbf{78.2} \\

\addlinespace\textit{XL} & & & & & & & \\
parallel &          81.9 &          79.4 &          71.0 &          84.5 &          80.4 &          70.2 &          77.4 \\
   +MASS &          82.4 &          81.0 & \textbf{76.2} &          85.1 &          82.6 & \textbf{79.0} & \textbf{80.8} \\
     +BT & \textbf{83.0} & \textbf{81.4} &          75.4 & \textbf{85.2} & \textbf{84.3} &          75.2 &          80.4 \\
\bottomrule
\end{tabular}

            \subcaption{COMET scores ($\uparrow$)}
            \label{tab:mmt-full-scale-ntrex-comet}
        \end{minipage}
        \caption{Results of all methods across different model scales evaluated on the \textbf{NTREX} (News) test set. The BT and DAE models have used the (Wikipedia) single-domain monolingual split.}
        \label{tab:mmt-full-scale-ntrex}
    \end{table*}

    \begin{table*}[bt]
        \centering
        \begin{minipage}{0.32\textwidth}
            \centering
            
\setlength{\tabcolsep}{1.5pt} %
\renewcommand{\arraystretch}{0.8} %
\scriptsize
\begin{tabular}{lrrrrrrr}
\toprule[1.5pt]
\multirow{2}[3]{*}{Model} &
\multicolumn{3}{c}{en$\rightarrow$xx} &
\multicolumn{3}{c}{xx$\rightarrow$en} &
\multicolumn{1}{c}{\multirow{2}[3]{*}{Mean}}
\\
\cmidrule(lr){2-4} \cmidrule(lr){5-7}
         &          High &           Med &           Low &          High &           Med & \multicolumn{2}{l}{Low} \\
\midrule
\addlinespace\textit{Base} & & & & & & & \\
parallel &          27.5 &          11.6 &          12.4 &          28.4 &          21.6 &          17.3 &          19.7 \\
   +MASS &          26.7 &          11.1 &          12.5 &          27.5 &          21.3 &          20.0 &          19.9 \\
     +BT & \textbf{30.2} & \textbf{14.9} & \textbf{17.2} & \textbf{32.3} & \textbf{26.5} & \textbf{24.0} & \textbf{24.2} \\

\addlinespace\textit{Big} & & & & & & & \\
parallel &          32.3 &          14.3 &          14.4 &          32.4 &          24.2 &          17.4 &          22.3 \\
   +MASS &          31.9 &          14.0 &          15.4 &          32.9 &          27.0 &          24.6 &          24.3 \\
     +BT & \textbf{34.5} & \textbf{18.4} & \textbf{19.8} & \textbf{36.8} & \textbf{32.2} & \textbf{28.7} & \textbf{28.3} \\

\addlinespace\textit{XL} & & & & & & & \\
parallel &          34.8 &          14.7 &          14.9 &          36.8 &          25.9 &          18.7 &          24.1 \\
   +MASS &          35.2 &          16.1 &          16.1 &          38.0 &          28.1 &          25.2 &          26.4 \\
     +BT & \textbf{38.0} & \textbf{20.7} & \textbf{21.5} & \textbf{41.0} & \textbf{36.8} & \textbf{32.9} & \textbf{31.7} \\
\bottomrule
\end{tabular}

            \subcaption{BLEU scores ($\uparrow$)}
            \label{tab:mmt-full-scale-tico19-bleu}
        \end{minipage}\hfill
        \begin{minipage}{0.32\textwidth}
            \centering
            
\setlength{\tabcolsep}{1.5pt} %
\renewcommand{\arraystretch}{0.8} %
\scriptsize
\begin{tabular}{lrrrrrrr}
\toprule[1.5pt]
\multirow{2}[3]{*}{Model} &
\multicolumn{3}{c}{en$\rightarrow$xx} &
\multicolumn{3}{c}{xx$\rightarrow$en} &
\multicolumn{1}{c}{\multirow{2}[3]{*}{Mean}}
\\
\cmidrule(lr){2-4} \cmidrule(lr){5-7}
         &          High &           Med &           Low &          High &           Med & \multicolumn{2}{l}{Low} \\
\midrule
\addlinespace\textit{Base} & & & & & & & \\
parallel &          49.4 &          41.2 &          42.9 &          57.8 &          50.2 &          45.1 &          47.7 \\
   +MASS &          48.6 &          39.7 &          41.4 &          56.6 &          49.5 &          48.1 &          47.3 \\
     +BT & \textbf{52.1} & \textbf{44.8} & \textbf{46.1} & \textbf{60.7} & \textbf{55.7} & \textbf{52.8} & \textbf{52.0} \\

\addlinespace\textit{Big} & & & & & & & \\
parallel &          53.3 &          45.5 &          46.8 &          61.0 &          52.4 &          45.4 &          50.6 \\
   +MASS &          52.8 &          44.8 &          48.0 &          61.0 &          55.2 &          53.3 &          52.5 \\
     +BT & \textbf{55.4} & \textbf{49.7} & \textbf{48.6} & \textbf{64.2} & \textbf{60.7} & \textbf{57.0} & \textbf{55.8} \\

\addlinespace\textit{XL} & & & & & & & \\
parallel &          55.1 &          45.5 &          47.4 &          64.2 &          53.7 &          47.2 &          52.1 \\
   +MASS &          55.1 &          47.2 & \textbf{49.5} &          64.7 &          56.1 &          55.2 &          54.7 \\
     +BT & \textbf{58.0} & \textbf{50.7} &          46.7 & \textbf{67.0} & \textbf{64.1} & \textbf{60.5} & \textbf{57.6} \\
\bottomrule
\end{tabular}

            \subcaption{chrF scores ($\uparrow$)}
            \label{tab:mmt-full-scale-tico19-chrf}
        \end{minipage}\hfill
        \begin{minipage}{0.32\textwidth}
            \centering
            
\setlength{\tabcolsep}{1.5pt} %
\renewcommand{\arraystretch}{0.8} %
\scriptsize
\begin{tabular}{lrrrrrrr}
\toprule[1.5pt]
\multirow{2}[3]{*}{Model} &
\multicolumn{3}{c}{en$\rightarrow$xx} &
\multicolumn{3}{c}{xx$\rightarrow$en} &
\multicolumn{1}{c}{\multirow{2}[3]{*}{Mean}}
\\
\cmidrule(lr){2-4} \cmidrule(lr){5-7}
         &          High &           Med &           Low &          High &           Med & \multicolumn{2}{l}{Low} \\
\midrule
\addlinespace\textit{Base} & & & & & & & \\
parallel &          75.6 &          70.8 &          65.7 &          80.5 &          76.0 &          70.5 &          72.9 \\
   +MASS &          74.0 &          68.8 &          64.1 &          79.7 &          75.6 &          74.2 &          72.6 \\
     +BT & \textbf{77.0} & \textbf{75.0} & \textbf{71.0} & \textbf{82.2} & \textbf{78.6} & \textbf{76.0} & \textbf{76.4} \\

\addlinespace\textit{Big} & & & & & & & \\
parallel &          80.3 &          76.4 &          69.9 &          83.4 &          79.2 &          73.1 &          76.7 \\
   +MASS &          79.9 &          75.6 &          70.9 &          83.6 &          81.4 & \textbf{80.5} &          78.5 \\
     +BT & \textbf{81.1} & \textbf{80.2} & \textbf{75.8} & \textbf{84.7} & \textbf{83.0} & \textbf{80.5} & \textbf{80.7} \\

\addlinespace\textit{XL} & & & & & & & \\
parallel &          82.4 &          76.8 &          70.4 &          85.5 &          80.9 &          75.7 &          78.3 \\
   +MASS &          82.7 &          78.3 &          72.7 &          85.9 &          83.2 &          83.2 &          80.8 \\
     +BT & \textbf{83.2} & \textbf{81.3} & \textbf{75.6} & \textbf{86.2} & \textbf{85.4} & \textbf{83.7} & \textbf{82.4} \\
\bottomrule
\end{tabular}

            \subcaption{COMET scores ($\uparrow$)}
            \label{tab:mmt-full-scale-tico19-comet}
        \end{minipage}
        \caption{Results of all methods across different model scales evaluated on the \textbf{TICO-19} (Medical) test set. The BT and DAE models have used the (Wikipedia) single-domain monolingual split.}
        \label{tab:mmt-full-scale-tico19}
    \end{table*}

    \newlength{\figspace}
    \setlength{\figspace}{10pt} %

    \begin{figure*}[t]
        \begin{minipage}[t]{0.3\textwidth}
            \centering
            \includegraphics[width=\linewidth]{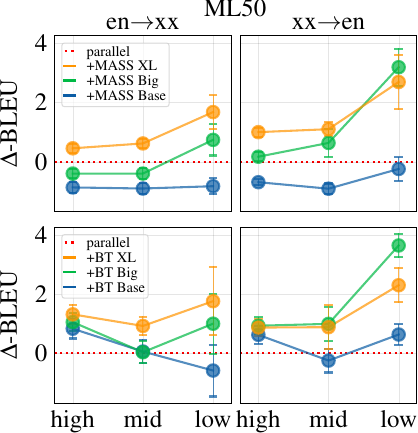}
            \par\vspace{\figspace}
            \includegraphics[width=\linewidth]{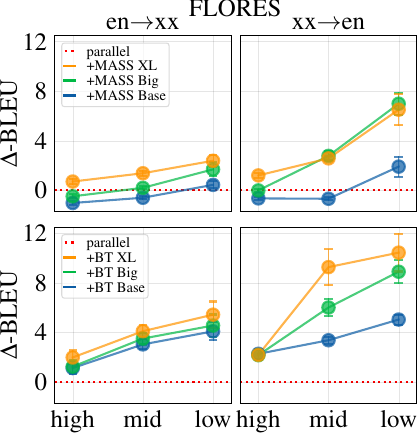}
            \par\vspace{\figspace}
            \includegraphics[width=\linewidth]{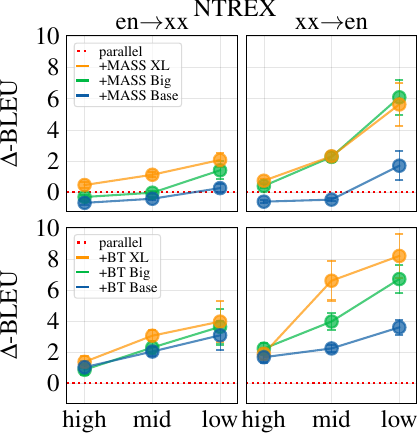}
            \par\vspace{\figspace}
            \includegraphics[width=\linewidth]{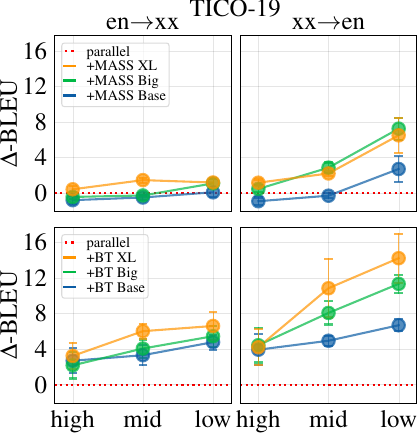}
            \caption{Mean BLEU differences (and standard error of the mean) per model with respect to the parallel-only baseline in the same scale (red dotted line).}
            \label{fig:app-scale-analysis-methods-bleu-detailed}
        \end{minipage}
        \hfill
        \begin{minipage}[t]{0.3\textwidth}
            \centering
            \includegraphics[width=\linewidth]{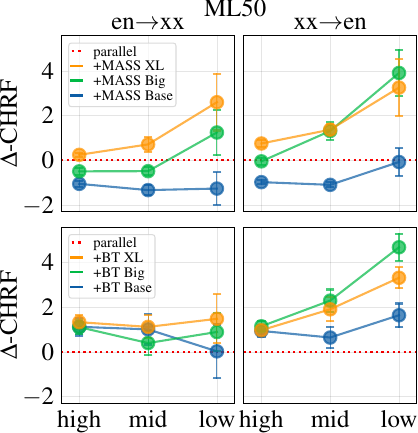}
            \par\vspace{\figspace}
            \includegraphics[width=\linewidth]{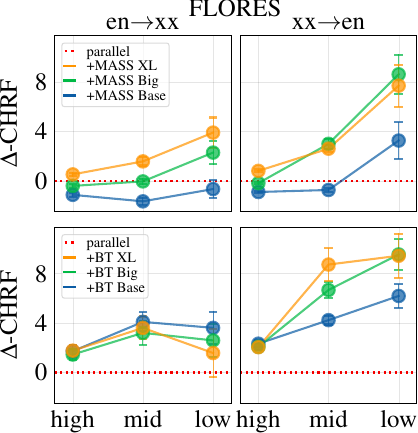}
            \par\vspace{\figspace}
            \includegraphics[width=\linewidth]{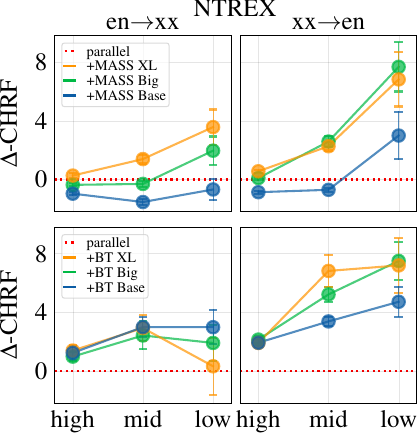}
            \par\vspace{\figspace}
            \includegraphics[width=\linewidth]{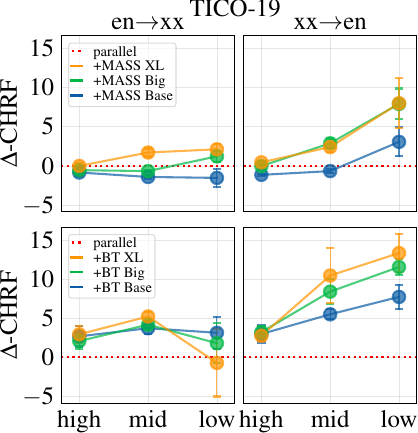}
            \caption{Mean chrF differences (and standard error of the mean) per model with respect to the parallel-only baseline in the same scale (red dotted line).}
            \label{fig:app-scale-analysis-methods-chrf-detailed}
        \end{minipage}
        \hfill
        \begin{minipage}[t]{0.3\textwidth}
            \centering
            \includegraphics[width=\linewidth]{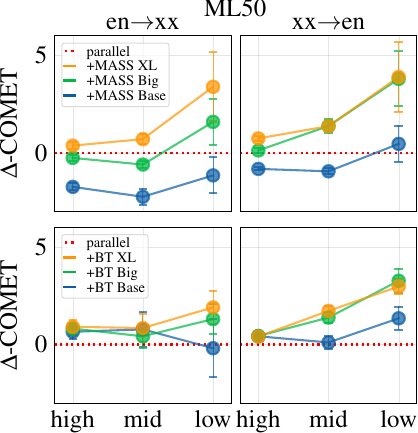}
            \par\vspace{\figspace}
            \includegraphics[width=\linewidth]{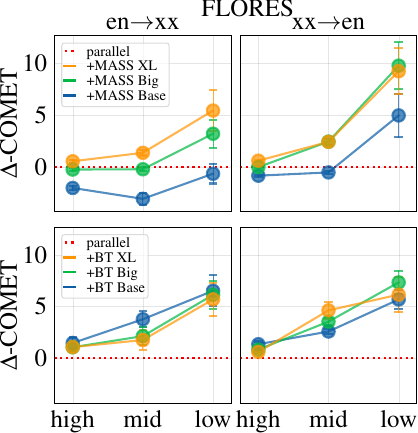}
            \par\vspace{\figspace}
            \includegraphics[width=\linewidth]{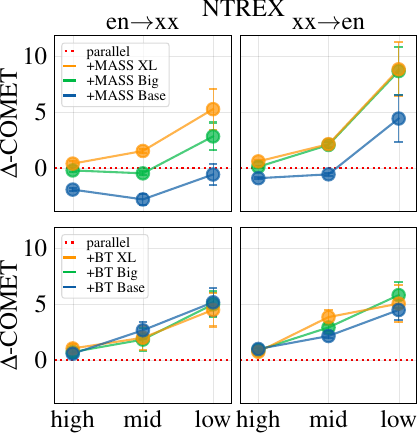}
            \par\vspace{\figspace}
            \includegraphics[width=\linewidth]{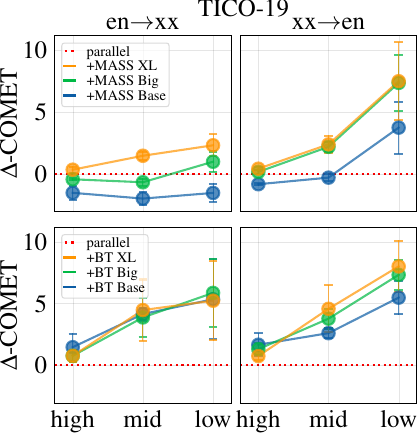}
            \caption{Mean COMET differences (and standard error of the mean) per model with respect to the parallel-only baseline in the same scale (red dotted line).}
            \label{fig:app-scale-analysis-methods-comet-detailed}
        \end{minipage}
    \end{figure*}

    \clearpage

    \section{Additional Tables and Figures}
    \definecolor{LightRed}{rgb}{1,0.8,0.8}

\begingroup
\setlength{\tabcolsep}{6pt} %
\renewcommand{\arraystretch}{1.1} %
\begin{table*}[!tb]
	\small
	\centering
	\begin{tabular}{ccrrrrr}
		\toprule[1.5pt]
		Group                     & Lang.    & Parallel   & Parallel + cap (10M) & wiki + cap (10M) & cc100 + cap (10M) & news + cap (10M) \\
		\midrule
		                          & cs       & 51,517,074 & 10,000,000           & 5,000,000        & 5,000,000         & 5,000,000        \\
		                          & de       & 45,992,835 & 10,000,000           & 5,000,000        & 5,000,000         & 5,000,000        \\
		                          & fr       & 38,507,539 & 10,000,000           & 5,000,000        & 5,000,000         & 5,000,000        \\
		                          & ja       & 17,203,227 & 10,000,000           & 5,000,000        & 5,000,000         & 5,000,000        \\
		                          & ru       & 13,599,766 & 10,000,000           & 5,000,000        & 5,000,000         & 5,000,000        \\
		                          & zh       & 11,173,646 & 10,000,000           & 5,000,000        & 5,000,000         & 5,000,000        \\
		                          & es       & 10,531,168 & 10,000,000           & 5,000,000        & 5,000,000         & 5,000,000        \\
		                          & pl       & 10,312,571 & 10,000,000           & 169,333          & 5,000,000         & 5,000,000        \\
		                          & lv       & 2,468,386  & 2,468,386            & 1,261,660        & 5,000,000         & 5,000,000        \\
		                          & fi       & 2,441,863  & 2,441,863            & 1,153,179        & 5,000,000         & 5,000,000        \\
		                          & hi       & 1,450,114  & 1,450,114            & 1,856,414        & 5,000,000         & 5,000,000        \\
		                          & lt       & 1,402,892  & 1,402,892            & 1,947,248        & 5,000,000         & 5,000,000        \\
		\rowcolor{LightRed}
		                          & iu       & 1,109,076  & 1,109,076            & *\textit{1,892}  & 0                 & 0                \\
		\multirow{-14}{*}{high}   & et       & 1,064,974  & 1,064,974            & 2,585,642        & 5,000,000         & 5,000,000        \\
		\midrule
		                          & ta       & 612,747    & 612,747              & 2,119,411        & 5,000,000         & 2,861,282        \\
		                          & ro       & 600,019    & 600,019              & 3,604,671        & 5,000,000         & 5,000,000        \\
		                          & si       & 594,438    & 594,438              & 443,711          & 5,000,000         & 0                \\
		                          & ps       & 573,218    & 573,218              & 391,604          & 2,000,879         & 1,096,628        \\
		                          & ne       & 504,085    & 504,085              & 328,219          & 5,000,000         & 0                \\
		                          & ml       & 343,668    & 343,668              & 1,481,937        & 5,000,000         & 1,423,835        \\
		                          & nl       & 232,038    & 232,038              & 5,000,000        & 5,000,000         & 2,967,745        \\
		                          & it       & 226,385    & 226,385              & 5,000,000        & 5,000,000         & 5,000,000        \\
		                          & ar       & 225,678    & 225,678              & 5,000,000        & 5,000,000         & 5,000,000        \\
		                          & ko       & 223,750    & 223,750              & 5,000,000        & 5,000,000         & 5,000,000        \\
		                          & he       & 204,468    & 204,468              & 5,000,000        & 5,000,000         & 0                \\
		                          & tr       & 203,702    & 203,702              & 5,000,000        & 5,000,000         & 5,000,000        \\
		                          & km       & 183,934    & 183,934              & 256,007          & 3,398,559         & 0                \\
		                          & fa       & 142,128    & 142,128              & 5,000,000        & 5,000,000         & 5,000,000        \\
		                          & vi       & 127,117    & 127,117              & 5,000,000        & 5,000,000         & 0                \\
		                          & hr       & 116,866    & 116,866              & 2,556,084        & 5,000,000         & 5,000,000        \\
		\multirow{-17}{*}{medium} & uk       & 104,021    & 104,021              & 5,000,000        & 5,000,000         & 2,222,071        \\
		\midrule
		                          & th       & 91,245     & 91,245               & 514,270          & 5,000,000         & 0                \\
		                          & id       & 83,932     & 83,932               & 5,000,000        & 5,000,000         & 2,378,340        \\
		                          & sv       & 53,580     & 53,580               & 5,000,000        & 5,000,000         & 0                \\
		                          & pt       & 49,431     & 49,431               & 5,000,000        & 5,000,000         & 5,000,000        \\
		                          & af       & 41,268     & 41,268               & 1,260,811        & 5,000,000         & 428,151          \\
		\rowcolor{LightRed}
		                          & xh       & 37,900     & 37,900               & *\textit{14,985} & 437,761           & 0                \\
		                          & kk       & 27,618     & 27,618               & 1,674,930        & 5,000,000         & 3,869,280        \\
		                          & ur       & 25,188     & 25,188               & 1,133,339        & 5,000,000         & 0                \\
		                          & mk       & 24,022     & 24,022               & 1,953,775        & 5,000,000         & 863,917          \\
		                          & te       & 21,513     & 21,513               & 1,568,018        & 5,000,000         & 3,461,218        \\
		                          & sl       & 18,714     & 18,714               & 2,340,732        & 5,000,000         & 0                \\
		                          & my       & 17,980     & 17,980               & 943,634          & 1,229,875         & 0                \\
		                          & ka       & 12,292     & 12,292               & 264,710          & 5,000,000         & 0                \\
		                          & gl       & 9,491      & 9,491                & 2,358,124        & 5,000,000         & 0                \\
		                          & mr       & 9,203      & 9,203                & 644,383          & 5,000,000         & 827,586          \\
		                          & mn       & 7,145      & 7,145                & 332,251          & 5,000,000         & 0                \\
		                          & gu       & 6,535      & 6,535                & 340,779          & 4,767,339         & 3,042,472        \\
		                          & az       & 5,652      & 5,652                & 2,355,880        & 5,000,000         & 0                \\
		\multirow{-19}{*}{low}    & bn       & 4,338      & 4,338                & 2,699,357        & 5,000,000         & 5,000,000        \\
		\bottomrule[1.5pt]
	\end{tabular}
	\caption{The statistics of the parallel and training data we use for each language. The red-highlighted rows show the languages that we remove from our experiments.}
	\label{table:full-data-stats}
\end{table*}
\endgroup

\end{document}